%% file: main.tex
\documentclass{article}

\usepackage{microtype}
\usepackage{graphicx}
\usepackage{subcaption}
\usepackage{booktabs} 

\usepackage{hyperref}

\usepackage[preprint]{icml2026}

\usepackage{amsmath}
\usepackage{amssymb}
\usepackage{mathtools}
\usepackage{amsthm}

\usepackage[capitalize,noabbrev]{cleveref}

\theoremstyle{plain}

\theoremstyle{definition}

\theoremstyle{remark}

\usepackage[textsize=tiny]{todonotes}

\icmltitlerunning{GLAIM: Adaptive Dependency Learning for Time Series Imputation}

\usepackage{amsfonts}
\usepackage{amsmath}
\usepackage{amssymb}
\usepackage{multirow}
\usepackage{makecell}

\begin{document}

\twocolumn[
  \icmltitle{GLAIM: Learning Global and Local Adaptive Inter-Variable Dependency \\ for Multivariate Time Series Imputation}

  \begin{icmlauthorlist}
    \icmlauthor{Mingyang Wang}{hnu}
    \icmlauthor{Rongwen Li}{hnu}
    \icmlauthor{Xiao Wang}{hnu}
    \icmlauthor{Changjian Chen}{hnu}
  \end{icmlauthorlist}

  \icmlaffiliation{hnu}{
    College of Computer Science and Electronic Engineering,
    Hunan University,
    Changsha,
    Hunan,
    China
  }
  \icmlcorrespondingauthor{Changjian Chen}{changjianchen@hnu.edu.cn}
  
  \vskip 0.3in
]

\printAffiliationsAndNotice{}  

\begin{abstract}
\input{Paper/0_Abstract}
\end{abstract}

\input{Paper/1_Introduction}
\input{Paper/2_Related_Work}
\input{Paper/3_Method}
\input{Paper/4_Experiments}
\input{Paper/5_Conclusion}

\bibliography{icml2026}
\bibliographystyle{icml2026}

\appendix
\newpage
\input{Paper/6_Appendix}

\end{document}

%% file: Paper/0_Abstract.tex
Multivariate time series imputation is fundamental to downstream analysis, yet modeling inter-variable dependencies with incomplete observations remains challenging.
Existing methods learn global dependencies across samples or dynamic local dependencies per sample.
Global dependencies are stable but adapt poorly to sample variations and temporal non-stationarity, whereas local dependencies are adaptive yet unreliable when observations are insufficient, causing erroneous information propagation.
To address these limitations, we propose GLAIM, a \textbf{G}lobal--\textbf{L}ocal \textbf{A}daptive \textbf{I}nter-variable Dependency \textbf{M}odeling framework for multivariate time series imputation.
GLAIM comprises two complementary components. The Stable Global Dependency Constructor derives robust global inter-variable dependencies from complementary temporal representations, providing a stable backbone less affected by sample-specific missingness and noise.
The Sample-Conditioned Dependency Refiner adapts this backbone to each sample and time step using its temporal state and available observations, enabling reliable local refinement under incomplete observations.
Extensive experiments on nine real-world datasets demonstrate that GLAIM achieves state-of-the-art performance under random and block missingness, remains robust to missing-rate shifts, and benefits from its complementary global and local components.

%% file: Paper/1_Introduction.tex
\section{Introduction}

Multivariate time series are prevalent in domains such as power systems~\cite{YYang2026a}, healthcare~\cite{HLi2025f}, finance~\cite{DCao2023k}, and meteorology~\cite{XZhao2025g}. 
In real-world applications, multivariate time series datasets often contain varying degrees of missingness due to factors such as sensor failures, communication disruptions, and omissions in manual records~\cite{QNater2025h, BAgbo2022d, KAggarwal2023m}.
Missing values can result in the loss of critical state information, thereby affecting the accuracy and reliability of downstream tasks such as forecasting, classification, and anomaly detection~\cite{JWang2025i, ZWang2024k, YZheng2024l}.
Therefore, imputing these missing values has become a fundamental problem in multivariate time series analysis.

\begin{figure}[t]
    \centering
    \includegraphics[width=0.95\columnwidth]{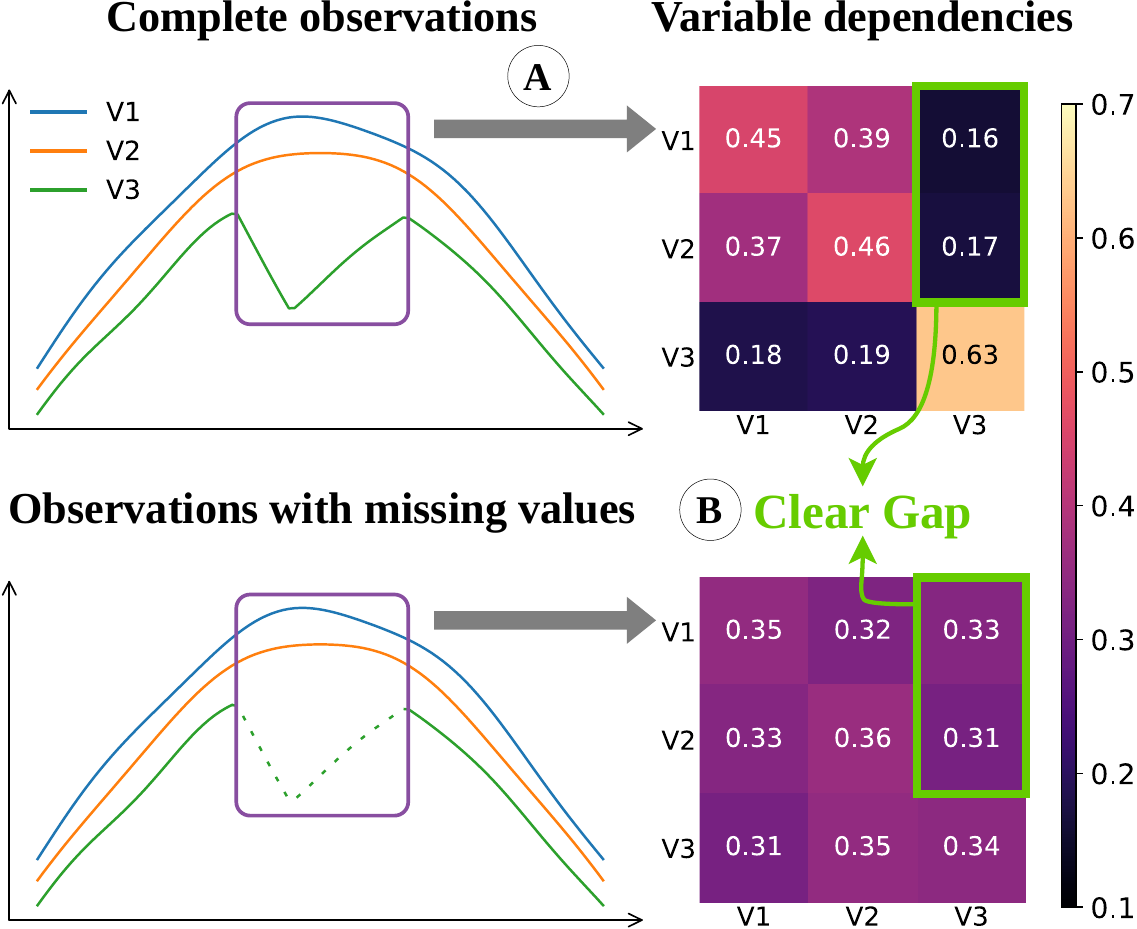} 
    \caption{Illustration of local dependencies estimated within a sliding window from complete observations and observations with missing values. Insufficient valid observations can make the estimated local dependencies unreliable.
}
\label{fig:intro}
\end{figure}

Existing time series imputation methods can be broadly categorized into intra- and inter-variable modeling methods based on whether dependencies are modeled within individual variables or across different variables. 
Compared to intra-variable modeling methods, inter-variable modeling methods additionally capture dependencies across variables, thereby achieving state-of-the-art imputation performance~\cite{MLiu2023c, ZLai2024a, ALaeeq2025a}.
These methods either learn global dependencies or estimate dynamic local dependencies specific to each sample.
Methods based on global dependency modeling capture stable inter-variable dependencies shared across samples but have limited adaptability to sample-level variations and temporal non-stationarity~\cite{ZWu2020b, CShang2021a, ACini2022a, TNie2024b}. 
In contrast, methods based on local dependency modeling adapt inter-variable dependencies to the sample-specific context (e.g., Figure~\ref{fig:intro}A) but require sufficient valid observations for reliable estimation~\cite{YZhang2023f, YLiu2024c, KObata2024d}. 
When valid observations are insufficient, the estimated local dependencies may become unreliable (e.g., Figure~\ref{fig:intro}B), resulting in erroneous information propagation~\cite{IMarisca2022b, GLiang2024e}.

Given the complementary strengths of global and local dependencies, a natural solution is to combine them. 
Such joint modeling can establish stable inter-variable dependencies shared across samples while adapting them to individual sample states. 
However, achieving this goal is non-trivial due to two challenges. 
\textbf{First}, learning global inter-variable dependencies shared across samples should be robust to sample-specific missingness patterns and noise, thereby establishing a reliable backbone for cross-variable information propagation. 
\textbf{Second}, building on this global dependency backbone, adapting inter-variable dependencies to individual samples should account for their temporal states and available observations while mitigating unreliable information propagation caused by incomplete observations.

To tackle these challenges, we propose GLAIM, a \textbf{G}lobal--\textbf{L}ocal \textbf{A}daptive \textbf{I}nter-variable Dependency \textbf{M}odeling framework for multivariate time series imputation.
The core of GLAIM consists of a Stable Global Dependency Constructor (StaGlo) and a Sample-Conditioned Dependency Refiner (SaCoRef).
StaGlo first captures complementary intra-variable temporal patterns to obtain informative variable representations, and then generates feature-wise global inter-variable dependencies from shared relation tokens via a parameterized generator. Since these dependencies are constructed independently of the current observations, StaGlo is less susceptible to sample-specific missingness patterns and noise.
SaCoRef summarizes each variable sequence into a variable-level token that jointly encodes its temporal state, observation pattern, and temporal order. 
It then models relations among these tokens to derive sample-specific relation context, which conditions time-specific cross-variable interactions and adaptively regulates information propagation, mitigating the adverse effects of incomplete observations.

Extensive experiments on nine real-world datasets demonstrate that GLAIM consistently achieves state-of-the-art imputation performance under both random and block missingness and remains robust to missing-rate shifts, while ablation studies and dependency visualizations further validate the complementary component contributions and the stability of the learned global dependencies.
We summarize our main contributions as follows:
\begin{itemize}
    \item We propose a unified global--local framework that combines stable dependencies shared across samples with adaptive dependencies conditioned on individual samples and time steps.
    \item We develop StaGlo for stable global dependency construction and SaCoRef for sample-conditioned local dependency refinement, improving the robustness of inter-variable modeling under incomplete observations.
    \item We conduct extensive experiments on nine real-world datasets, demonstrating state-of-the-art imputation performance and strong robustness under diverse missingness settings.
\end{itemize}

%% file: Paper/2_Related_Work.tex
\section{Related Work}

\subsection{Intra-Variable Temporal Modeling}
Intra-variable temporal modeling recovers missing observations by exploiting the temporal dynamics of individual variables. 
Deterministic methods commonly use pattern-based estimation, localized data selection, or meta-learning to directly estimate missing values \cite{Darji2024Binned, Almeida2025Meta}.
For example, TS-Pothole \cite{Sanwouo2024TSPothole} exploits recurring temporal patterns through recursive imputation, while WBDI \cite{Phan2024WBDI} combines forward and backward estimates using weighted bidirectional imputation.

Probabilistic methods instead model the conditional distribution of missing values through latent-variable, diffusion, continuous-function, or state-space formulations \cite{JMiguel2023b, ELNaour2024g}. 
Representative approaches include BGaP \cite{HMAhmed2022e}, which performs Bayesian Gaussian-process imputation with temporal forcing, and OLVGP \cite{WWang2024h}, which models temporal uncertainty through online variational Gaussian-process inference.
Although these methods effectively capture temporal dynamics, independently modeling variables limits their ability to exploit complementary information from correlated variables.

\subsection{Inter-Variable Dependency Modeling}
Inter-variable dependency modeling uses correlated variables to provide complementary information for missing-value estimation. Existing methods generally learn either globally shared dependencies or sample-dependent local dependencies.

Global dependency methods encode stable variable relationships as graph-structured, spectral, or low-rank priors \cite{DCao2020d, CShang2021a, KYi2023i}. 
MTGNN \cite{ZWu2020b} and GRIN \cite{ACini2022a} both represent variables as graph nodes and model inter-variable dependencies through information propagation over a shared graph structure. ImputeFormer \cite{TNie2024b} introduces a low-rank structural prior to capture stable shared patterns in multivariate spatio-temporal data. 
Such shared structures provide stable relational information but are less responsive to sample-level variations and temporal non-stationarity.

Local dependency methods condition variable relationships on the input sample using mechanisms such as dynamic graphs, conditional priors, or channel grouping \cite{YLiu2022c, MLiu2023c, XQiu2025c}. 
iTransformer \cite{YLiu2024c} applies attention across variable-wise sequence tokens, while MissNet \cite{KObata2024d} and TimeFilter \cite{YHu2025d} model dynamic relationships through regime-switching networks and temporal relation filtering, respectively. 
These methods improve adaptability, but dependency estimation from incomplete observations may become unreliable when valid values are sparse.

Unlike existing methods that rely solely on either globally shared or sample-dependent dependencies, GLAIM first constructs stable global dependencies with StaGlo and then refines them for individual samples with SaCoRef.

%% file: Paper/3_Method.tex
\section{Method}

\begin{figure*}[t]
\centering
\includegraphics[width=0.95\textwidth]{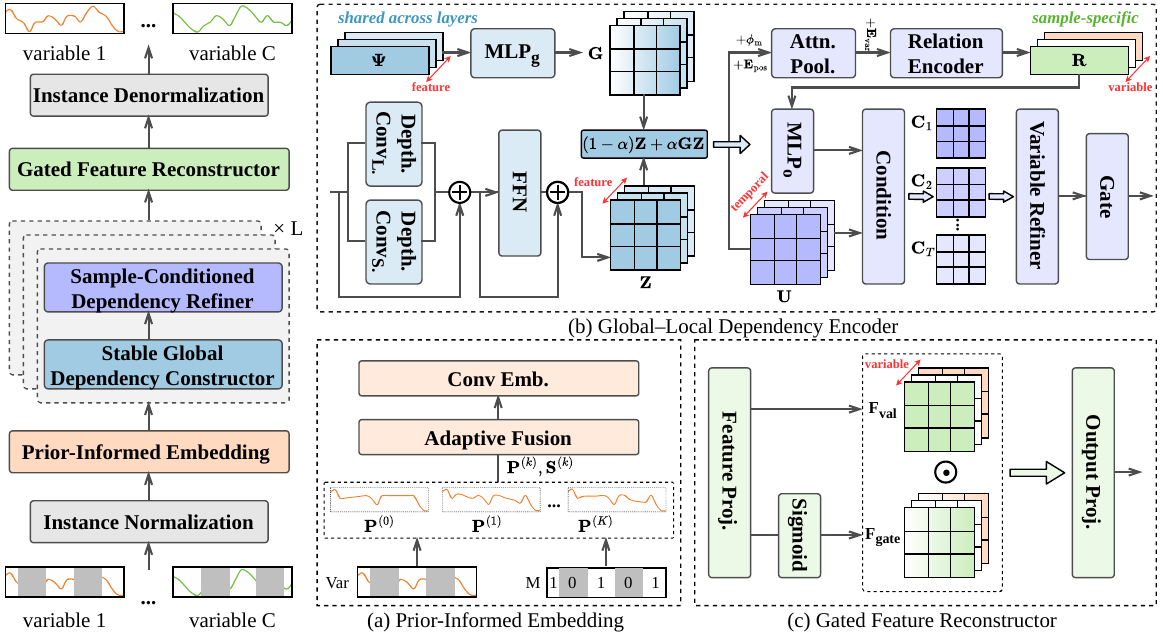}
\caption{Overview of GLAIM for multivariate time series imputation. The model consists of (a) Prior-Informed Embedding (PriEmb), which adaptively fuses multi-scale temporal priors and embeds them with the observation mask; (b) the Global--Local Dependency Encoder (GloLoc), comprising $L$ stacked pairs of the Stable Global Dependency Constructor (StaGlo) and Sample-Conditioned Dependency Refiner (SaCoRef); and (c) the Gated Feature Reconstructor (GateRecon), which decodes the learned representations before instance denormalization.}
\label{fig:method}
\end{figure*}

\subsection{Problem Definition}
Given a multivariate time series $\mathbf{X} \in \mathbb{R}^{C \times T}$, where $C$ denotes the number of variables and $T$ denotes the number of time steps, each entry $x_{ct}$ represents the value of the $c$-th variable at the $t$-th time step.
To characterize missingness, we define a binary observation mask $\mathbf{M} \in \{0,1\}^{C \times T}$, where $m_{ct}=1$ indicates that $x_{ct}$ is observed and $m_{ct}=0$ indicates that it is missing.
The observed time series is given by $\mathbf{X}^{\mathrm{obs}} = \mathbf{X} \odot \mathbf{M}$. The objective of time series imputation is to estimate the missing values through:
\begin{equation}
    \hat{\mathbf{X}} = f_\theta(\mathbf{X}^{\mathrm{obs}}, \mathbf{M}),
\end{equation}
where $\hat{\mathbf{X}} \in \mathbb{R}^{C \times T}$ denotes the imputed time series.

\subsection{Overall Architecture}
As illustrated in Figure~\ref{fig:method}, GLAIM comprises three main components: the \textbf{Pri}or-Informed \textbf{Emb}edding (PriEmb), the \textbf{Glo}bal--\textbf{Loc}al Dependency Encoder (GloLoc), and the \textbf{Gate}d Feature \textbf{Recon}structor (GateRecon).
Specifically, PriEmb constructs a mask-aware temporal prior $\mathbf{P} \in \mathbb{R}^{C \times T}$ and projects it together with the observation mask into an initial representation $\mathbf{H}^{(0)} \in \mathbb{R}^{C \times D \times T}$, where $D$ denotes the hidden dimension.
GloLoc then produces a representation with encoded dependencies, $\mathbf{H}^{(L)} \in \mathbb{R}^{C \times D \times T}$, through $L$ stacked layers that construct stable global inter-variable dependencies and subsequently refine sample-adaptive local inter-variable dependencies.
Finally, GateRecon decodes $\mathbf{H}^{(L)}$ to obtain the final reconstruction $\hat{\mathbf{X}}$.

\paragraph{Prior-Informed Embedding.}
As illustrated in Figure~\ref{fig:method}(a), PriEmb constructs a mask-aware intra-variable temporal prior $\mathbf{P}$ that provides a temporally coherent initialization at missing positions.
Before constructing the prior, PriEmb normalizes each variable using statistics computed from its observed values, yielding $\widetilde{\mathbf{X}}^{\mathrm{obs}} \in \mathbb{R}^{C \times T}$.
It then constructs a linear interpolation candidate $\mathbf{P}^{(0)} \in \mathbb{R}^{C \times T}$ from neighboring observations, with a support map $\mathbf{S}^{(0)} \in \mathbb{R}^{C \times T}$ determined by the distances to the nearest observations on both sides.
PriEmb further generates multi-scale smoothing candidates $\{\mathbf{P}^{(k)}\}_{k=1}^{K}$ by applying average pooling to the observations using different window sizes. Each support map $\mathbf{S}^{(k)}$ is defined by the proportion of observed values within the corresponding window.
Smaller windows capture local variations, whereas larger windows yield smoother trends.
Detailed formulations are provided in the Appendix~\ref{priemb}.

To adapt the temporal scale to varying missingness patterns, PriEmb predicts position-wise fusion weights from the interpolation candidate, the observation mask, and the candidate-specific support maps:
\begin{equation}
    \boldsymbol{\Omega} = \operatorname{Router}\left(\mathbf{P}^{(0)}, \mathbf{M}, \{\mathbf{S}^{(k)}\}_{k=0}^K\right),
\end{equation}
where $\boldsymbol{\Omega}=\{\boldsymbol{\Omega}^{(k)}\}_{k=0}^{K}$, and each $\boldsymbol{\Omega}^{(k)} \in \mathbb{R}^{C \times T}$ denotes the contribution of the $k$-th candidate.
The adaptively fused candidate is computed as:
\begin{equation}
    \overline{\mathbf{P}} = \sum_{k=0}^{K} \boldsymbol{\Omega}^{(k)} \odot \mathbf{P}^{(k)}.
\end{equation}
PriEmb retains normalized observations and fills missing positions with the fused candidate:
\begin{equation}
    \mathbf{P} = \mathbf{M} \odot \widetilde{\mathbf{X}}^{\mathrm{obs}} + (\mathbf{1}-\mathbf{M}) \odot \overline{\mathbf{P}}.
\end{equation}

After obtaining the temporal prior, PriEmb stacks it with the observation mask to form a two-channel tensor $\mathbf{X}_{\mathrm{in}} = \operatorname{Stack}\left(\mathbf{P},\mathbf{M}\right) \in \mathbb{R}^{C \times 2 \times T}$.
This tensor is projected using a 1D convolution with a kernel size of 1:
\begin{equation}
    \mathbf{H}^{(0)} = \operatorname{Conv1D}\left(\mathbf{X}_{\mathrm{in}}\right) \in \mathbb{R}^{C \times D \times T}.
\end{equation}
Acting only along channels, the projection fuses prior values and missingness into a unified latent representation while preserving temporal resolution.

\paragraph{Global--Local Dependency Encoder.}
GloLoc integrates temporal and inter-variable information through $L$ stacked layers.
Each layer combines stable global dependency construction with sample-conditioned local dependency refinement.
The detailed design of GloLoc is presented in the following subsection.

\paragraph{Gated Feature Reconstructor.}
As illustrated in Figure~\ref{fig:method}(c), GateRecon decodes the representation with encoded dependencies, $\mathbf{H}^{(L)}$, through a gated latent-feature transformation.
It first projects the hidden representation into a value branch and a gate branch:
\begin{equation}
    \left[\mathbf{F}_{\mathrm{val}},\mathbf{F}_{\mathrm{gate}}\right] = \operatorname{FeatureProj}\left(\operatorname{Norm}\left(\mathbf{H}^{(L)}\right)\right),
\end{equation}
where $\mathbf{F}_{\mathrm{val}}$ and $\mathbf{F}_{\mathrm{gate}}$ have the same shape and represent the candidate reconstruction features and their feature-wise gates, respectively.
The normalized reconstruction is produced by sigmoid gating and output projection:
\begin{equation}
    \widetilde{\mathbf{X}}^{\mathrm{rec}} = \operatorname{OutProj}\left(\mathbf{F}_{\mathrm{val}} \odot \sigma\left(\mathbf{F}_{\mathrm{gate}}\right)\right) \in \mathbb{R}^{C \times T}.
\end{equation}
The observed values are preserved, whereas the decoded values are used at missing positions:
\begin{equation}
    \widetilde{\mathbf{X}} = \mathbf{M} \odot \widetilde{\mathbf{X}}^{\mathrm{obs}} + (\mathbf{1}-\mathbf{M}) \odot \widetilde{\mathbf{X}}^{\mathrm{rec}}.
\end{equation}
Finally, $\widetilde{\mathbf{X}}$ is transformed back to the original data scale to produce $\hat{\mathbf{X}}$.

During training, the reconstruction loss is defined as the mean squared error (MSE) over the entries indicated as missing by $\mathbf{M}$:
\begin{equation}
    \mathcal{L}
    =
    \frac{
        \left\|
            (\mathbf{1}-\mathbf{M})
            \odot
            (\hat{\mathbf{X}}-\mathbf{X})
        \right\|_F^2
    }{
        \left\|\mathbf{1}-\mathbf{M}\right\|_1
    }.
\end{equation}

\subsection{Global--Local Dependency Encoder}
As illustrated in Figure~\ref{fig:method}(b), GloLoc consists of $L$ stacked layers, each containing a \textbf{Sta}ble \textbf{Glo}bal Dependency Constructor (StaGlo) and a \textbf{Sa}mple-\textbf{Co}nditioned Dependency \textbf{Ref}iner (SaCoRef).
Specifically, in the $l$-th layer, StaGlo first extracts complementary intra-variable temporal patterns and then constructs observation-independent global inter-variable dependencies from shared relation tokens.
SaCoRef subsequently derives sample-specific relation context and uses it to condition local inter-variable refinement at each time step.
The computation in the $l$-th layer is formulated as:
\begin{equation}
    \begin{aligned}
        \mathbf{U}^{(l)}
            &= \operatorname{StaGlo}^{(l)}\left(
                \mathbf{H}^{(l-1)}
            \right),\\
        \mathbf{H}^{(l)}
            &= \operatorname{SaCoRef}^{(l)}\left(
                \mathbf{U}^{(l)}, \mathbf{M}
            \right),
    \end{aligned}
\end{equation}
where $\mathbf{U}^{(l)}$ and $\mathbf{H}^{(l)}$ denote the representations after stable global dependency construction and sample-conditioned local dependency refinement, respectively, both in $\mathbb{R}^{C \times D \times T}$.

\paragraph{Stable Global Dependency Constructor.}
Reliable global dependency construction requires temporally informative variable representations without premature cross-variable propagation from incomplete observations.
StaGlo therefore performs stable global dependency construction in four stages: it enhances each variable with dual-scale temporal convolutions, refines the resulting temporal features, generates observation-independent feature-wise global dependencies from shared relation tokens, and gates global variable aggregation. This ordering establishes informative variable representations before any cross-variable information is propagated.

Given the input $\mathbf{H}^{(l-1)} \in \mathbb{R}^{C \times D \times T}$, StaGlo first applies two parallel depthwise temporal convolutions independently to each variable.
The large-kernel branch captures broader temporal patterns, whereas the small-kernel branch preserves fine-grained local variations.
Because the convolutions operate independently along the temporal dimension, they introduce no cross-variable interaction at this stage. Their outputs are aggregated through a residual connection:
\begin{equation}
\overline{\mathbf{Z}}^{(l)}
    = \mathbf{H}^{(l-1)}
    + \sum_{j \in \{\mathrm{L},\mathrm{S}\}}
        \operatorname{DWConv}_{j}^{(l)}\left(
            \operatorname{Norm}\left(\mathbf{H}^{(l-1)}\right)
        \right),
\end{equation}
where $\overline{\mathbf{Z}}^{(l)} \in \mathbb{R}^{C \times D \times T}$ is the dual-scale temporal representation.
A feed-forward network then refines these temporal features through a second residual connection:
\begin{equation}
    \mathbf{Z}^{(l)}
        = \overline{\mathbf{Z}}^{(l)}
        + \operatorname{FFN}^{(l)}\left(
            \operatorname{Norm}\left(
                \overline{\mathbf{Z}}^{(l)}
            \right)
          \right),
\end{equation}
where $\mathbf{Z}^{(l)} \in \mathbb{R}^{C \times D \times T}$ provides temporally informative variable representations for subsequent global dependency construction without premature information propagation across variables.

StaGlo next uses learnable global relation tokens $\boldsymbol{\Psi}=\{\boldsymbol{\psi}_d \in \mathbb{R}^{d_\psi}\}_{d=1}^{D}$ shared across all GloLoc layers.
The shared tokens encode global relation prototypes, while independent generators allow each layer to construct dependencies suited to its representation level. For each feature dimension, the layer-specific generator maps the corresponding token to two variable-side representation matrices:
\begin{equation}
    \left[\mathbf{A}_{d}^{(l)},\mathbf{B}_{d}^{(l)}\right]
    = \operatorname{MLP}_{g}^{(l)}\left(\boldsymbol{\psi}_d\right),
\end{equation}
where $\mathbf{A}_{d}^{(l)},\mathbf{B}_{d}^{(l)} \in \mathbb{R}^{C \times d_{\mathrm{g}}}$.
The feature-wise global dependency matrix is constructed as:
\begin{equation}
    \mathbf{G}_{d}^{(l)} = \operatorname{Softmax}\left(
        \mathbf{A}_{d}^{(l)}{\mathbf{B}_{d}^{(l)}}^{\top}
        \big/
        \sqrt{d_{\mathrm{g}}}
    \right),
\end{equation}
where $\mathbf{G}_{d}^{(l)} \in \mathbb{R}^{C \times C}$.
Because it is generated from shared relation tokens and layer-specific parameters rather than current observations, the matrix is less affected by sample-specific missingness patterns and noise, while different feature dimensions can represent distinct variable relations.

Let $\mathbf{Z}_{d}^{(l)} \in \mathbb{R}^{C \times T}$ denote the temporal representation for the $d$-th feature dimension across variables and time steps.
StaGlo then aggregates information across variables using the corresponding global dependency matrix:
\begin{equation}
    \overline{\mathbf{U}}_{d}^{(l)} = \mathbf{G}_{d}^{(l)}\mathbf{Z}_{d}^{(l)}.
\end{equation}
Finally, a learnable feature-wise gate regulates global variable mixing:
\begin{equation}
    \mathbf{U}_{d}^{(l)}
        = \left(1-\alpha_{d}^{(l)}\right)\mathbf{Z}_{d}^{(l)}
        + \alpha_{d}^{(l)}
          \overline{\mathbf{U}}_{d}^{(l)},
\end{equation}
where $\alpha_{d}^{(l)} \in (0,1)$ controls the contribution of global aggregation for the $d$-th feature dimension, balancing the original temporal representation and globally propagated information.
The updated feature dimensions are recombined into the stable globally encoded representation $\mathbf{U}^{(l)}$.

\input{Tables/random}

\paragraph{Sample-Conditioned Dependency Refiner.}
StaGlo captures stable inter-variable dependencies shared across samples but cannot fully reflect the temporal states and observation patterns of individual samples.
SaCoRef therefore performs sample-conditioned refinement in four stages: it augments each variable sequence with missingness and temporal-order information, derives sample-specific relation context with a RelationEncoder, uses this context to condition time-specific cross-variable interactions, and applies a feature-wise dynamic gate to regulate the resulting contextual increments.

Let $\mathbf{U}_{c}^{(l)} \in \mathbb{R}^{T \times D}$ denote the temporal representation of variable $c$, and let $\mathbf{m}_{c} \in \{0,1\}^{T \times 1}$ denote its observation mask.
SaCoRef first augments the sequence with missingness and temporal-order information:
\begin{equation}
    \mathbf{Y}_{c}^{(l)}
        =\mathbf{U}_{c}^{(l)}
        + \phi_{\mathrm{m}}^{(l)}(\mathbf{m}_{c})
        + \mathbf{E}_{\mathrm{pos}}^{(l)},
\end{equation}
where $\phi_{\mathrm{m}}^{(l)}(\cdot)$ is the mask embedding and
$\mathbf{E}_{\mathrm{pos}}^{(l)} \in \mathbb{R}^{T \times D}$ is a learnable position embedding.
The augmented sequence is summarized into a variable-level token through temporal attention pooling:
\begin{equation}
    \mathbf{v}_{c}^{(l)}
        =
        \operatorname{AttnPool}_{t}^{(l)}
        \left(\mathbf{Y}_{c}^{(l)}\right)
        \in \mathbb{R}^{D},
\end{equation}
where $\mathbf{v}_{c}^{(l)}$ encodes the temporal state, observation pattern, and temporal order of variable $c$. This pooling adaptively emphasizes informative temporal positions when forming the variable-level summary.

The variable-level tokens are stacked as $\mathbf{V}^{(l)} \in \mathbb{R}^{C \times D}$.
SaCoRef then adds a learnable variable identity embedding and models their relations:
\begin{equation}
    \begin{aligned}
        \mathbf{R}^{(l)}
        =
        \operatorname{RelationEncoder}^{(l)}
        \left(
            \mathbf{V}^{(l)}
            + \mathbf{E}_{\mathrm{var}}^{(l)}
        \right),
    \end{aligned}
\end{equation}
where $\mathbf{E}_{\mathrm{var}}^{(l)} \in \mathbb{R}^{C \times D}$ distinguishes the variables.
The RelationEncoder applies self-attention across variables, producing sample-specific relation context $\mathbf{R}^{(l)} \in \mathbb{R}^{C \times D}$ that reflects their heterogeneous observation patterns.

The relation context is next mapped to feature-wise scale and shift parameters:
\begin{equation}
    [
        \mathbf{R}_{\mathrm{s}}^{(l)},
        \mathbf{R}_{\mathrm{b}}^{(l)}
    ]
    =
    \operatorname{MLP}_{o}^{(l)}
    (\mathbf{R}^{(l)}).
\end{equation}
Let $\mathbf{U}_{t}^{(l)} \in \mathbb{R}^{C \times D}$ denote the representations of all variables at time $t$.
The relation-conditioned representation is:
\begin{equation}
    \begin{aligned}
        \mathbf{C}_{t}^{(l)}
        ={}& \mathbf{U}_{t}^{(l)}
        + \operatorname{Linear}_{\mathrm{out}}^{(l)}
        \big(
            \operatorname{Linear}_{\mathrm{in}}^{(l)}
            (\mathbf{U}_{t}^{(l)}) \\
        &\odot
            \tanh(\mathbf{R}_{\mathrm{s}}^{(l)})
        \big)
        + \mathbf{R}_{\mathrm{b}}^{(l)}.
    \end{aligned}
\end{equation}
The scale and shift parameters are shared across time steps within each sample, providing consistent sample-level conditioning while preserving the time-specific states carried by $\mathbf{U}_{t}^{(l)}$ before cross-variable interaction.

SaCoRef then applies self-attention across variables at each time step and extracts the resulting contextual increment:
\begin{equation}
    \begin{aligned}
        \boldsymbol{\Delta}_{t}^{(l)}
        ={}&
        \operatorname{VariableRefiner}^{(l)}
        (\mathbf{C}_{t}^{(l)})
        - \mathbf{C}_{t}^{(l)},
    \end{aligned}
\end{equation}
where $\operatorname{VariableRefiner}^{(l)}$ is a transformer encoder operating along the variable dimension.
By isolating the contextual increment introduced by cross-variable interaction, SaCoRef can selectively incorporate this information into the globally encoded representation rather than directly adopting the refined representation.
Finally, a feature-wise dynamic gate regulates this increment:
\begin{equation}
\begin{aligned}
    \mathbf{J}_{t}^{(l)}
    &=
    \sigma\!\left(
        \operatorname{Gate}^{(l)}
        \left(
            [
                \mathbf{U}_{t}^{(l)}
                \Vert
                \mathbf{R}^{(l)}
            ]
        \right)
    \right), \\
    \mathbf{H}_{t}^{(l)}
    &=
    \mathbf{U}_{t}^{(l)}
    +
    \mathbf{J}_{t}^{(l)}
    \odot
    \boldsymbol{\Delta}_{t}^{(l)}.
\end{aligned}
\end{equation}
Here, $\mathbf{J}_{t}^{(l)} \in (0,1)^{C \times D}$ depends jointly on the current variable states and the sample-specific relation context, allowing SaCoRef to suppress potentially unreliable contextual increments under incomplete observations.
Stacking the updated representations over time yields $\mathbf{H}^{(l)}$.

%% file: Tables/random.tex
\begin{table*}[t]
\centering
\small
\setlength{\tabcolsep}{1mm}
\caption{Imputation performance under random missingness, averaged across missing rates of $0.1$, $0.3$, $0.5$, and $0.7$. The best and second-best results are highlighted in bold and underlined, respectively.}
\begin{tabular}{c|cc|cc|cc|cc|cc|cc|cc|cc}
\toprule
\multirow{2}{*}{Dataset} & \multicolumn{2}{c|}{\textbf{GLAIM (Ours)}} & \multicolumn{2}{c|}{ModernTCN} & \multicolumn{2}{c|}{iTransformer} & \multicolumn{2}{c|}{TimesNet} & \multicolumn{2}{c|}{PatchTST} & \multicolumn{2}{c|}{ImputeFormer} & \multicolumn{2}{c|}{SAITS} & \multicolumn{2}{c}{PSW-I} \\
 & MSE & MAE & MSE & MAE & MSE & MAE & MSE & MAE & MSE & MAE & MSE & MAE & MSE & MAE & MSE & MAE \\
\midrule
ETTh1 & \textcolor{red}{\textbf{0.048}} & \textcolor{red}{\textbf{0.139}} & {\underline{\textcolor{blue}{0.063}}} & 0.165 & 0.137 & 0.257 & 0.093 & 0.201 & 0.111 & 0.215 & 0.100 & 0.191 & 0.066 & {\underline{\textcolor{blue}{0.163}}} & 0.116 & 0.208 \\
ETTh2 & \textcolor{red}{\textbf{0.038}} & \textcolor{red}{\textbf{0.115}} & {\underline{\textcolor{blue}{0.048}}} & {\underline{\textcolor{blue}{0.140}}} & 0.124 & 0.240 & 0.053 & 0.150 & 0.066 & 0.161 & 0.105 & 0.200 & 0.123 & 0.224 & 0.059 & 0.148 \\
ETTm1 & \textcolor{red}{\textbf{0.021}} & \textcolor{red}{\textbf{0.090}} & 0.026 & 0.102 & 0.072 & 0.181 & 0.034 & 0.119 & 0.047 & 0.136 & 0.025 & 0.101 & {\underline{\textcolor{blue}{0.024}}} & {\underline{\textcolor{blue}{0.099}}} & 0.044 & 0.130 \\
ETTm2 & \textcolor{red}{\textbf{0.018}} & \textcolor{red}{\textbf{0.070}} & {\underline{\textcolor{blue}{0.022}}} & {\underline{\textcolor{blue}{0.088}}} & 0.061 & 0.161 & 0.024 & 0.094 & 0.031 & 0.102 & 0.036 & 0.105 & 0.043 & 0.135 & 0.031 & 0.097 \\
Weather & \textcolor{red}{\textbf{0.027}} & \textcolor{red}{\textbf{0.039}} & {\underline{\textcolor{blue}{0.029}}} & {\underline{\textcolor{blue}{0.046}}} & 0.043 & 0.087 & 0.033 & 0.060 & 0.034 & 0.052 & 0.040 & 0.074 & 0.041 & 0.092 & 0.034 & 0.046 \\
PEMS03 & \textcolor{red}{\textbf{0.017}} & \textcolor{red}{\textbf{0.083}} & 0.031 & 0.119 & 0.042 & 0.139 & 0.041 & 0.139 & 0.042 & 0.144 & {\underline{\textcolor{blue}{0.031}}} & {\underline{\textcolor{blue}{0.113}}} & 0.063 & 0.149 & 0.041 & 0.141 \\
Exchange & \textcolor{red}{\textbf{0.002}} & \textcolor{red}{\textbf{0.023}} & 0.003 & 0.032 & 0.017 & 0.085 & 0.004 & 0.035 & 0.003 & 0.031 & 0.023 & 0.088 & 0.068 & 0.179 & {\underline{\textcolor{blue}{0.003}}} & {\underline{\textcolor{blue}{0.029}}} \\
Illness & \textcolor{red}{\textbf{0.051}} & \textcolor{red}{\textbf{0.110}} & 0.236 & 0.293 & 0.726 & 0.560 & {\underline{\textcolor{blue}{0.209}}} & 0.281 & 0.246 & 0.284 & 0.936 & 0.552 & 0.490 & 0.390 & 0.346 & {\underline{\textcolor{blue}{0.250}}} \\
Electricity & \textcolor{red}{\textbf{0.045}} & \textcolor{red}{\textbf{0.133}} & 0.079 & 0.192 & 0.086 & 0.198 & 0.103 & 0.222 & 0.073 & 0.182 & {\underline{\textcolor{blue}{0.053}}} & {\underline{\textcolor{blue}{0.148}}} & 0.171 & 0.283 & 0.065 & 0.168 \\
\midrule
Avg & \textcolor{red}{\textbf{0.030}} & \textcolor{red}{\textbf{0.089}} & {\underline{\textcolor{blue}{0.060}}} & {\underline{\textcolor{blue}{0.131}}} & 0.145 & 0.212 & 0.066 & 0.145 & 0.072 & 0.145 & 0.150 & 0.175 & 0.121 & 0.190 & 0.082 & 0.135 \\
\bottomrule
\end{tabular}
\label{tab:random}
\end{table*}

%% file: Paper/4_Experiments.tex
\section{Experiments}

We conduct comprehensive experiments to evaluate GLAIM under diverse missingness conditions. The evaluation covers standard random and block missing patterns, robustness to missing-rate shifts, visualization of learned global dependencies, and ablation studies. Together, these experiments assess imputation accuracy, generalization, structural stability, and the contribution of each model component.

\subsection{Experimental Setup}

\paragraph{Datasets.}
We conduct experiments on nine real-world multivariate time series datasets: ETTh1, ETTh2, ETTm1, and ETTm2~\cite{ETT} for electricity transformer measurements; Electricity~\cite{ECL} for electricity consumption; Weather~\cite{WeatherILI} for meteorological observations; Illness~\cite{WeatherILI} for influenza-like illness statistics; Exchange~\cite{Exchange} for exchange rates; and PEMS03~\cite{PEMS} for road traffic. 

\paragraph{Baselines.}
We compare GLAIM with general-purpose time series models (ModernTCN~\cite{ModernTCN}, iTransformer~\cite{YLiu2024c}, TimesNet~\cite{TimesNet}, PatchTST~\cite{PatchTST}) and specialized imputation methods (ImputeFormer~\cite{TNie2024b}, SAITS~\cite{WDu2023h}, and PSW-I~\cite{PSWI}).

\paragraph{Implementation Details.}
The input window length is 96. GLAIM is implemented in PyTorch~\cite{PyTroch} and trained with Adam~\cite{Adam}, a batch size of 16, and at most 300 epochs. 
We apply early stopping with a patience of 30 epochs based on validation masked MSE and use the best checkpoint for testing. General-purpose baselines use the same training settings, whereas specialized methods follow their original protocols. MSE and MAE are computed only at artificially masked positions in the normalized space. Experiments run on NVIDIA RTX 4090 GPUs.

\input{Tables/block}

\subsection{Main Results}

\paragraph{Random Missing.}
We randomly mask individual observations at missing rates of $0.1$, $0.3$, $0.5$, and $0.7$.
Table~\ref{tab:random} reports results averaged over the four rates. GLAIM ranks first in all 18 dataset--metric comparisons and achieves average MSE and MAE values of $0.030$ and $0.089$. Compared with the strongest overall competitor, ModernTCN, it reduces average MSE and MAE by $50.0\%$ and $32.1\%$, respectively. 
The consistent improvements across all nine datasets indicate that GLAIM is effective across diverse domains and temporal characteristics rather than benefiting only a few datasets. This uniform ranking avoids dataset-specific trade-offs across both metrics.

\paragraph{Block Missing.}
We randomly remove contiguous segments from each variable until missing rates of $0.1$, $0.3$, $0.5$, and $0.7$ are reached.
As shown in Table~\ref{tab:block}, GLAIM again ranks first in all 18 dataset--metric comparisons, with average MSE and MAE values of $0.050$ and $0.113$. 
Relative to ModernTCN, the average errors decrease by $55.0\%$ and $33.9\%$, respectively. The gains remain substantial on difficult datasets such as Illness, PEMS03, and Exchange, while GLAIM also preserves an advantage on Electricity, where ImputeFormer is competitive. 
GLAIM therefore recovers long gaps using evidence outside each segment. These results demonstrate robust performance under both isolated and contiguous missing observations.

\paragraph{Missing-Rate Shift.}
\input{Tables/robust}
We further test robustness to missing-rate shifts by training trainable models at a block missing rate of $0.3$ and evaluating them at rates of $0.1$, $0.3$, $0.5$, and $0.7$. 
Table~\ref{tab:robust} reports MSE averaged over all nine datasets; PSW-I is excluded because it is training-free. 
Although errors increase with the test missing rate, GLAIM achieves the best MSE at every rate, including unseen rates. Its MSE rises from $0.029$ at $0.1$ to $0.124$ at $0.7$, while its reduction over the strongest baselines remains above $50\%$ even at the highest rate. The gradual degradation suggests that learned dependencies remain informative as observations become increasingly sparse.

Overall, the results support the global--local dependency modeling strategy of GLAIM. Observation-independent global dependencies provide shared cross-variable structure that is less sensitive to missingness shifts, whereas sample-conditioned refinement adapts variable interactions to the temporal state and available evidence of each sample. Their combination yields stable advantages across missingness patterns and rates.

\subsection{Dependency Visualization}
To examine the stability of the global dependencies learned by StaGlo, we separately train GLAIM on ETTh1 under random missing rates of $0.1$, $0.3$, $0.5$, and $0.7$, while keeping all other settings unchanged. This controlled setting isolates the influence of missingness severity on the learned cross-variable structure.
We then visualize the aggregated dependency matrices produced by the first StaGlo layer, enabling a direct comparison of whether the dominant dependency patterns change as observations become increasingly sparse.

\begin{figure}[t]
    \centering
    \includegraphics[width=0.85\columnwidth]{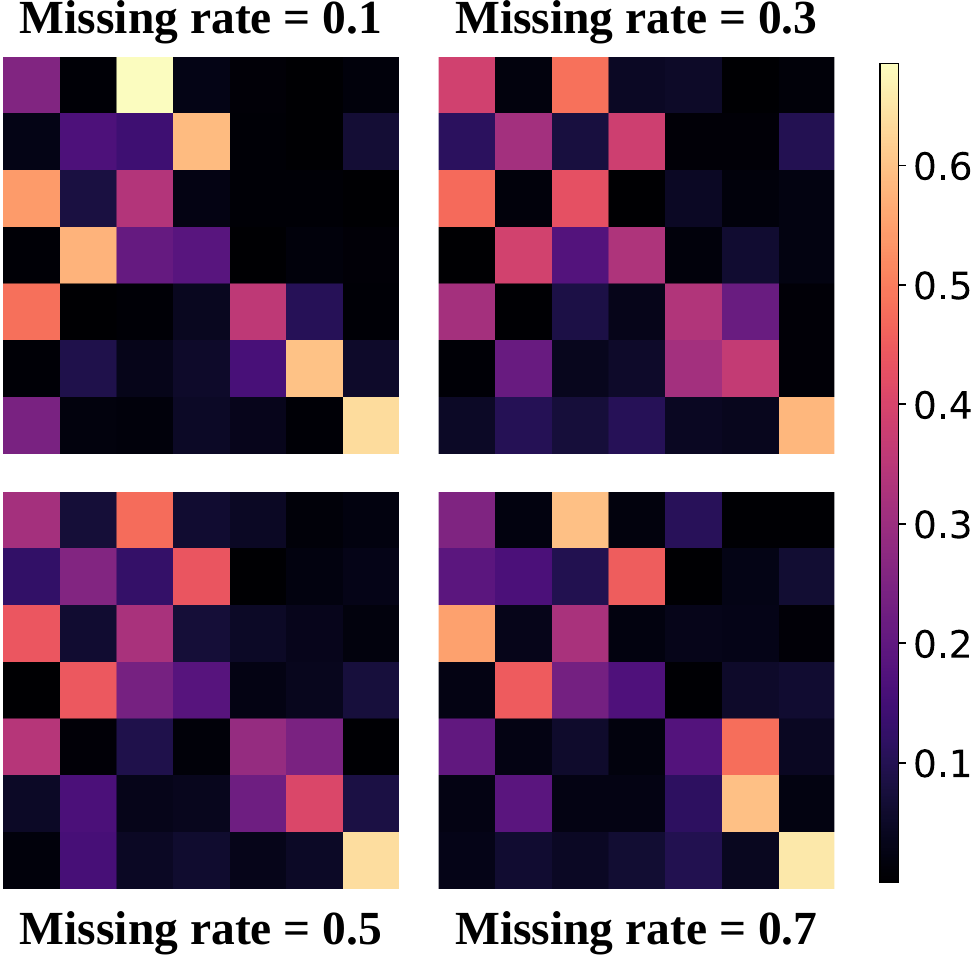}
    \caption{Global dependency matrices from the first StaGlo layer on ETTh1, showing visually consistent structures across missing rates.}
    \label{fig:exp}
\end{figure}

As shown in Figure~\ref{fig:exp}, the dependency matrices exhibit visually consistent structures across different missing rates. 
To further assess this consistency, we evaluate both the rank agreement and the distributional discrepancy between each pair of matrices. 
Across all pairwise comparisons, the matrices show consistently high Spearman correlations ($0.767$--$0.892$) and low Jensen--Shannon divergences ($0.029$--$0.074$). This consistency remains evident even between the extreme missing rates of $0.1$ and $0.7$.

These results indicate that StaGlo preserves both the relative ordering and the overall distribution of inter-variable dependencies under varying missingness conditions, supporting its ability to capture stable global dependency structures.

\subsection{Ablation Study}
We conduct ablation studies on six datasets. Models are trained at a block missing rate of $0.3$ and evaluated at rates of $0.1$, $0.3$, and $0.5$; Table~\ref{tab:ablation} reports MSE averaged over these rates. The full model achieves the lowest MSE.

\input{Tables/ablation}

\paragraph{Dependency Modeling.}
Removing StaGlo causes the largest degradation, increasing MSE to $0.077$ ($67.4\%$), which confirms that its stable global dependency backbone is essential for cross-variable information propagation.
This substantial gap also suggests that sample-conditioned refinement cannot recover reliable interactions when the structural prior is absent.
Removing SaCoRef increases MSE to $0.049$ ($6.5\%$), showing that fixed global dependencies alone cannot fully capture sample-specific temporal states and missingness patterns.
Thus, StaGlo and SaCoRef play complementary roles: the former provides shared structure, while the latter adaptively calibrates interactions for each sample.

\paragraph{Initialization and Reconstruction.}
Replacing PriEmb with the raw masked input increases MSE to $0.052$ ($13.0\%$), validating the mask-aware temporal prior as a coherent initialization for missing positions.
This larger degradation suggests that reconstruction quality depends strongly on how incomplete sequences are represented before subsequent processing.
Removing the gated transformation in GateRecon yields an MSE of $0.049$ ($6.5\%$), indicating that adaptive modulation of candidate features is more effective than direct output projection.
Together, these results verify the complementary contributions of prior-informed initialization, global and sample-conditioned dependency modeling, and gated reconstruction.

%% file: Tables/block.tex
\begin{table*}[t]
\centering
\small
\setlength{\tabcolsep}{1mm}
\caption{Imputation performance under block missingness, averaged across missing rates of $0.1$, $0.3$, $0.5$, and $0.7$. The best and second-best results are highlighted in bold and underlined, respectively.}
\begin{tabular}{c|cc|cc|cc|cc|cc|cc|cc|cc}
\toprule
\multirow{2}{*}{Dataset} & \multicolumn{2}{c|}{\textbf{GLAIM (Ours)}} & \multicolumn{2}{c|}{ModernTCN} & \multicolumn{2}{c|}{iTransformer} & \multicolumn{2}{c|}{TimesNet} & \multicolumn{2}{c|}{PatchTST} & \multicolumn{2}{c|}{ImputeFormer} & \multicolumn{2}{c|}{SAITS} & \multicolumn{2}{c}{PSW-I} \\
 & MSE & MAE & MSE & MAE & MSE & MAE & MSE & MAE & MSE & MAE & MSE & MAE & MSE & MAE & MSE & MAE \\
\midrule
ETTh1 & \textcolor{red}{\textbf{0.076}} & \textcolor{red}{\textbf{0.171}} & 0.113 & 0.217 & 0.228 & 0.323 & 0.175 & 0.268 & 0.193 & 0.283 & 0.192 & 0.266 & {\underline{\textcolor{blue}{0.109}}} & {\underline{\textcolor{blue}{0.194}}} & 0.809 & 0.516 \\
ETTh2 & \textcolor{red}{\textbf{0.049}} & \textcolor{red}{\textbf{0.136}} & {\underline{\textcolor{blue}{0.067}}} & {\underline{\textcolor{blue}{0.173}}} & 0.141 & 0.259 & 0.079 & 0.186 & 0.095 & 0.201 & 0.175 & 0.266 & 0.199 & 0.299 & 0.293 & 0.346 \\
ETTm1 & \textcolor{red}{\textbf{0.031}} & \textcolor{red}{\textbf{0.108}} & 0.046 & 0.136 & 0.124 & 0.235 & 0.072 & 0.166 & 0.086 & 0.184 & 0.047 & 0.137 & {\underline{\textcolor{blue}{0.039}}} & {\underline{\textcolor{blue}{0.123}}} & 0.996 & 0.579 \\
ETTm2 & \textcolor{red}{\textbf{0.024}} & \textcolor{red}{\textbf{0.085}} & {\underline{\textcolor{blue}{0.032}}} & {\underline{\textcolor{blue}{0.110}}} & 0.074 & 0.180 & 0.036 & 0.118 & 0.050 & 0.135 & 0.058 & 0.139 & 0.069 & 0.168 & 0.215 & 0.281 \\
Weather & \textcolor{red}{\textbf{0.036}} & \textcolor{red}{\textbf{0.049}} & {\underline{\textcolor{blue}{0.041}}} & {\underline{\textcolor{blue}{0.063}}} & 0.065 & 0.108 & 0.049 & 0.078 & 0.052 & 0.072 & 0.051 & 0.085 & 0.058 & 0.113 & 0.166 & 0.182 \\
PEMS03 & \textcolor{red}{\textbf{0.019}} & \textcolor{red}{\textbf{0.086}} & 0.039 & 0.132 & 0.051 & 0.149 & 0.060 & 0.169 & 0.052 & 0.155 & {\underline{\textcolor{blue}{0.030}}} & {\underline{\textcolor{blue}{0.107}}} & 0.065 & 0.151 & 0.399 & 0.398 \\
Exchange & \textcolor{red}{\textbf{0.004}} & \textcolor{red}{\textbf{0.035}} & 0.007 & 0.052 & 0.014 & 0.077 & 0.008 & 0.055 & {\underline{\textcolor{blue}{0.006}}} & {\underline{\textcolor{blue}{0.047}}} & 0.033 & 0.110 & 0.139 & 0.299 & 0.022 & 0.091 \\
Illness & \textcolor{red}{\textbf{0.140}} & \textcolor{red}{\textbf{0.189}} & 0.547 & 0.441 & 0.982 & 0.626 & {\underline{\textcolor{blue}{0.510}}} & {\underline{\textcolor{blue}{0.419}}} & 0.674 & 0.472 & 1.276 & 0.649 & 0.676 & 0.474 & 1.868 & 0.770 \\
Electricity & \textcolor{red}{\textbf{0.065}} & \textcolor{red}{\textbf{0.160}} & 0.104 & 0.220 & 0.117 & 0.229 & 0.131 & 0.253 & 0.104 & 0.212 & {\underline{\textcolor{blue}{0.066}}} & {\underline{\textcolor{blue}{0.163}}} & 0.197 & 0.304 & 0.170 & 0.266 \\
\midrule
Avg & \textcolor{red}{\textbf{0.050}} & \textcolor{red}{\textbf{0.113}} & {\underline{\textcolor{blue}{0.111}}} & {\underline{\textcolor{blue}{0.171}}} & 0.199 & 0.243 & 0.125 & 0.190 & 0.146 & 0.196 & 0.214 & 0.214 & 0.172 & 0.236 & 0.549 & 0.381 \\
\bottomrule
\end{tabular}
\label{tab:block}
\end{table*}

%% file: Tables/robust.tex
\begin{table}[t]
\centering
\caption{Robustness to missing-rate shifts under block missingness. All trainable models are trained at a missing rate of $0.3$ and evaluated at missing rates of $0.1$, $0.3$, $0.5$, and $0.7$. The MSE values are averaged across all nine datasets.}
\begin{tabular}{lcccc}
\toprule
\multicolumn{1}{l}{Model} & \multicolumn{1}{c}{0.1} & \multicolumn{1}{c}{0.3} & \multicolumn{1}{c}{0.5} & \multicolumn{1}{c}{0.7} \\
\midrule
ModernTCN & {\underline{\textcolor{blue}{0.073}}} & {\underline{\textcolor{blue}{0.091}}} & 0.147 & 0.267 \\
iTransformer & 0.209 & 0.184 & 0.442 & 1.161 \\
TimesNet & 0.088 & 0.101 & {\underline{\textcolor{blue}{0.146}}} & {\underline{\textcolor{blue}{0.251}}} \\
PatchTST & 0.103 & 0.138 & 0.184 & 0.275 \\
ImputeFormer & 0.131 & 0.178 & 0.255 & 0.368 \\
SAITS & 0.122 & 0.155 & 0.222 & 0.367 \\
\midrule
\textbf{GLAIM (Ours)} & \textcolor{red}{\textbf{0.029}} & \textcolor{red}{\textbf{0.039}} & \textcolor{red}{\textbf{0.062}} & \textcolor{red}{\textbf{0.124}} \\
\bottomrule
\end{tabular}
\label{tab:robust}
\end{table}

%% file: Tables/ablation.tex
\begin{table}[t]
\centering
\setlength{\tabcolsep}{1.5mm}
\caption{Ablation results under block missingness on six datasets. All variants are trained at a missing rate of $0.3$ and evaluated at missing rates of $0.1$, $0.3$, and $0.5$. The reported MSE values are averaged across the three evaluation rates.}
\begin{tabular}{cccccc}
\toprule
Model & \textbf{\makecell{GLAIM \\ (Ours)}} & \makecell{w/o \\ StaGlo} & \makecell{w/o \\ SaCoRef} & \makecell{w/o \\ Prior} & \makecell{w/o \\ Gated} \\
\midrule
ETTh1 & \textbf{0.061} & 0.111 & 0.067 & 0.066 & 0.070 \\
ETTh2 & \textbf{0.044} & 0.073 & 0.046 & 0.044 & 0.045 \\
ETTm1 & \textbf{0.026} & 0.033 & 0.027 & 0.028 & 0.028 \\
ETTm2 & \textbf{0.021} & 0.029 & 0.023 & 0.022 & 0.021 \\
Exchange & \textbf{0.004} & 0.005 & 0.004 & 0.004 & 0.004 \\
Illness & \textbf{0.118} & 0.215 & 0.127 & 0.150 & 0.129 \\
\midrule
Avg & \textbf{0.046} & 0.077 & 0.049 & 0.052 & 0.049 \\
\midrule
$\Delta (\%) \downarrow$ & -- & +67.4 & +6.5 & +13.0 & +6.5 \\
\bottomrule
\end{tabular}
\label{tab:ablation}
\end{table}

%% file: Paper/5_Conclusion.tex
\section{Conclusion}

We proposed GLAIM, a global--local adaptive inter-variable dependency modeling framework for multivariate time series imputation. 
StaGlo constructs a robust global dependency backbone from complementary temporal representations, capturing stable cross-variable structures shared across samples while reducing sensitivity to sample-specific missingness and noise. 
Building on this backbone, SaCoRef conditions dependency refinement on temporal states and available observations, enabling reliable adaptation of variable interactions to individual samples and time steps.
Experiments on nine real-world datasets demonstrate that GLAIM consistently achieves state-of-the-art performance under random and block missing settings and remains robust to missing-rate shifts. Ablation studies and dependency analyses further confirm the complementary roles of StaGlo and SaCoRef and the stability of the learned global dependencies.
Future work will explore extending GLAIM to more complex missingness patterns and broader multivariate time series tasks.

%% file: Paper/6_Appendix.tex
\section{Prior-Informed Embedding}
\label{priemb}
This section provides the detailed formulation of the temporal prior construction used in Prior-Informed Embedding (PriEmb). We adopt variable-first notation, omit the batch dimension, and let
$\widetilde{\mathbf{X}}^{\mathrm{obs}}\in\mathbb{R}^{C\times T}$ and
$\mathbf{M}\in\{0,1\}^{C\times T}$
denote the normalized observations and the corresponding binary observation mask, respectively. Before constructing the temporal prior, each variable is normalized using statistics computed only from its observed entries.

\paragraph{Linear interpolation candidate.}
For variable $c$ and time step $t$, let
\begin{equation}
    \begin{aligned}
        \ell_{ct}
        &=
        \max\{s\leq t:m_{cs}=1\},\\
        \qquad
        r_{ct}
        &=
        \min\{s\geq t:m_{cs}=1\}
    \end{aligned}
\end{equation}
denote the nearest observed time steps at or to the left and right of $t$, respectively.
The linear interpolation candidate is defined as:
\begin{equation}
    P_{ct}^{(0)}
    =
    \frac{r_{ct}-t}{r_{ct}-\ell_{ct}}
    \widetilde{X}^{\mathrm{obs}}_{c\ell_{ct}}
    +
    \frac{t-\ell_{ct}}{r_{ct}-\ell_{ct}}
    \widetilde{X}^{\mathrm{obs}}_{cr_{ct}}.
    \label{eq:appendix_interpolation}
\end{equation}
To expose the reliability of the neighboring observations to the router, we additionally construct two normalized distance maps:
\begin{equation}
    D_{ct}^{\mathrm{L}}
    =
    \min\left(
        \frac{t-\ell_{ct}}{w_{\max}},1
    \right),
    \quad
    D_{ct}^{\mathrm{R}}
    =
    \min\left(
        \frac{r_{ct}-t}{w_{\max}},1
    \right),
    \label{eq:appendix_interpolation_distance}
\end{equation}
where $w_{\max}$ is the largest smoothing-window size.

\paragraph{Mask-weighted multi-scale smoothing.}
Let $\{w_k\}_{k=1}^{K}$ be a set of odd-valued window sizes, and let
$\mathcal{N}_{k}(t)$ denote the temporal neighborhood centered at $t$ with width $w_k$.
For each scale, the smoothing candidate is computed using only observed entries:
\begin{equation}
    P_{ct}^{(k)}
    =
    \frac{
        \sum_{s\in\mathcal{N}_{k}(t)}
        m_{cs}\widetilde{X}^{\mathrm{obs}}_{cs}
    }{
        \sum_{s\in\mathcal{N}_{k}(t)}
        m_{cs}
    },
    \qquad k=1,\ldots,K.
    \label{eq:appendix_smoothing}
\end{equation}
Its support map is the proportion of observed values within the corresponding window:
\begin{equation}
    S_{ct}^{(k)}
    =
    \frac{1}{w_k}
    \sum_{s\in\mathcal{N}_{k}(t)}m_{cs}.
    \label{eq:appendix_smoothing_support}
\end{equation}
Thus, smaller windows preserve short-term variations, whereas larger windows provide smoother temporal estimates.

\paragraph{Adaptive candidate fusion.}
For each variable, the router input is formed by concatenating the interpolation candidate, the observation mask, the two directional distance maps, and the support maps of all smoothing candidates:
\begin{equation}
    \mathbf{Q}
    =
    \operatorname{Concat}\left(
        \mathbf{P}^{(0)},
        \mathbf{M},
        \mathbf{D}^{\mathrm{L}},
        \mathbf{D}^{\mathrm{R}},
        \mathbf{S}^{(1)},\ldots,\mathbf{S}^{(K)}
    \right),
    \label{eq:appendix_router_input}
\end{equation}
where the concatenation is performed along the candidate/support channel and $\mathbf{Q}\in\mathbb{R}^{C\times(K+4)\times T}$. The router consists of a temporal convolution, a GELU activation, and a pointwise output projection:
\begin{equation}
    \mathbf{A}
    =
    \operatorname{Conv}_{1\times1}
    \left(
        \operatorname{GELU}
        \left(
            \operatorname{Conv}_{1\times3}(\mathbf{Q})
        \right)
    \right)
    \in\mathbb{R}^{C\times(K+1)\times T}.
    \label{eq:appendix_router}
\end{equation}
The candidate weights are normalized along the candidate dimension using a temperature-scaled softmax:
\begin{equation}
    \Omega_{ct}^{(k)}
    =
    \frac{
        \exp\left(A_{ct}^{(k)}/\tau\right)
    }{
        \sum_{j=0}^{K}
        \exp\left(A_{ct}^{(j)}/\tau\right)
    },
    \qquad k=0,\ldots,K,
    \label{eq:appendix_router_weight}
\end{equation}
where $\tau$ denotes the routing temperature. The candidate estimates are then combined using the normalized routing weights:
\begin{equation}
    \overline{\mathbf{P}}
    =
    \sum_{k=0}^{K}
    \boldsymbol{\Omega}^{(k)}
    \odot
    \mathbf{P}^{(k)}.
    \label{eq:appendix_prior_fusion}
\end{equation}
The final temporal prior retains the normalized observations at observed positions and uses the fused candidate only at missing positions:
\begin{equation}
    \mathbf{P}
    =
    \mathbf{M}\odot\widetilde{\mathbf{X}}^{\mathrm{obs}}
    +
    (\mathbf{1}-\mathbf{M})\odot\overline{\mathbf{P}}.
    \label{eq:appendix_prior}
\end{equation}

\paragraph{Mask-aware input projection.}
The temporal prior and observation mask are stacked as two input channels:
\begin{equation}
    \mathbf{X}_{\mathrm{in}}
    =
    \operatorname{Stack}(\mathbf{P},\mathbf{M})
    \in\mathbb{R}^{C\times2\times T},
\end{equation}
and projected into the hidden space through a $1\times1$ convolution:
\begin{equation}
    \mathbf{H}^{(0)}
    =
    \operatorname{Conv}_{1\times1}
    \left(\mathbf{X}_{\mathrm{in}}\right)
    \in\mathbb{R}^{C\times D\times T}.
    \label{eq:appendix_prior_embedding}
\end{equation}
This projection jointly encodes the prior value and its observation status without changing the temporal resolution.

For compact notation, the support information is written as
$\{\mathbf{S}^{(k)}\}_{k=0}^{K}$.
In the implementation, the interpolation support $\mathbf{S}^{(0)}$ is represented by the two directional distance maps
$\mathbf{D}^{\mathrm{L}}$ and $\mathbf{D}^{\mathrm{R}}$, whereas
$\mathbf{S}^{(k)}$ for $k\geq1$ corresponds directly to the local observation density in Eq.~\eqref{eq:appendix_smoothing_support}.

\section{Implementation Details}

\subsection{Dataset Details}
\input{Appendix/Tables/dataset_statistics}
We evaluate GLAIM on nine real-world multivariate time series datasets covering electricity, meteorology, public health, finance, and transportation. Table~\ref{tab:dataset_statistics} summarizes their dimensionality, chronological splits, sampling frequencies, and time ranges.

\paragraph{ETT.}
ETTh1, ETTh2, ETTm1, and ETTm2~\cite{ETT} contain seven electricity-transformer variables. The hourly and 15-minute variants provide two temporal resolutions for evaluating imputation under electricity-system dynamics.

\paragraph{Electricity.}
Electricity (ECL)~\cite{ECL} records hourly electricity consumption from 321 variables. Its high dimensionality is useful for assessing large-scale multivariate imputation.

\paragraph{Weather.}
Weather~\cite{WeatherILI} contains 21 meteorological variables sampled every 10 minutes, representing densely observed environmental time series.

\paragraph{Illness.}
Illness (ILI)~\cite{WeatherILI} contains seven weekly influenza-like illness variables. Its relatively short and low-frequency sequence provides a distinct public-health setting.

\paragraph{Exchange.}
Exchange~\cite{Exchange} contains eight daily exchange-rate variables spanning more than two decades, representing long-term financial dynamics.

\paragraph{PEMS03.}
PEMS03~\cite{PEMS} contains 358 road-traffic variables sampled every five minutes. It provides a high-dimensional setting with fine-grained temporal variation.

\subsection{Experiment Details}

\paragraph{GLAIM configuration.}
Unless stated otherwise, GLAIM uses the same configuration across all datasets. The input window length is fixed to 96, and the hidden dimension $D$ is set to 128. The Global--Local Dependency Encoder (GloLoc) contains $L=2$ stacked layers, each comprising a Stable Global Dependency Constructor (StaGlo) followed by a Sample-Conditioned Dependency Refiner (SaCoRef). The feed-forward dimension is 256, the number of attention heads is 8, and the dropout rate is 0.1.

In StaGlo, the two parallel depthwise temporal convolutions use kernel sizes 5 and 71 to capture fine-grained and broad intra-variable temporal patterns, respectively. 
The shared global relation tokens $\boldsymbol{\Psi}$ have dimension $d_{\psi}=136$. 
The variable-side representation dimension used to construct the feature-wise global dependency matrices, $d_{\mathrm{g}}$, is set to 8 by default. It is increased to 32 for Electricity and 36 for PEMS03 to accommodate their larger numbers of variables. 
In SaCoRef, both the RelationEncoder operating on variable-level tokens and the VariableRefiner operating across variables at each time step contain one transformer encoder layer in every GloLoc layer.
For the Prior-Informed Embedding (PriEmb), the multi-scale smoothing candidates are constructed using average-pooling window sizes $\{3,5,9,17,33\}$. The Router has a hidden dimension of 16 and uses a temperature of 1.0.

\paragraph{Experiment Design.}
We evaluate GLAIM under random and block missingness at missing rates of $0.1$, $0.3$, $0.5$, and $0.7$. For the rate-specific evaluations, a separate model is trained for every combination of dataset, missingness pattern, and missing rate, ensuring that each model is optimized for the corresponding corruption setting. The masking patterns are applied independently to each input window. Under random missingness, each entry is independently selected for artificial masking with probability equal to the prescribed missing rate. Under block missingness, each variable is masked independently: the required number of masked time points is first determined by the missing rate and then divided into multiple nonempty contiguous segments. Their lengths and locations are randomly sampled, with at least one observed time point retained between adjacent segments. This construction preserves the target missing rate while producing irregular and temporally separated gaps within each variable.

During training and validation, artificial masks are generated dynamically, exposing the model to different missing positions throughout optimization. In contrast, the test masks are generated deterministically and cached before evaluation. Specifically, the mask assigned to each test sample is determined by the dataset, missingness pattern, missing rate, input-window length, and sample index. 
Due to the substantial computational cost of evaluating all dataset, missingness-pattern, and missing-rate combinations, each result is obtained from a single run using a fixed random seed of 2021.

To assess robustness to missing-rate shifts, we train models under block missingness at a fixed rate of $0.3$ and evaluate the resulting checkpoints without further adaptation at test missing rates of $0.1$, $0.3$, $0.5$, and $0.7$. The rate of $0.3$ therefore represents the matched setting, whereas the remaining rates measure generalization to milder or more severe missingness than encountered during training. We additionally report the corresponding shift experiment under random missingness using the same training and test rates.

\paragraph{Evaluation metrics.}
We report mean squared error (MSE) and mean absolute error (MAE) in the normalized space. Both metrics are computed exclusively over the set of artificially masked positions, denoted by $\mathcal{H}$, where each $(c,t)\in\mathcal{H}$ identifies an entry of variable $c$ at time step $t$ that is deliberately masked in the model input while its ground-truth value is retained for evaluation. The metrics are defined as follows:
\begin{equation}
    \begin{aligned}
        \mathrm{MSE}
        &=\frac{1}{|\mathcal{H}|}\sum_{(c,t)\in\mathcal{H}}
        (\hat{X}_{ct}-X_{ct})^2,\\
        \mathrm{MAE}
        &=\frac{1}{|\mathcal{H}|}\sum_{(c,t)\in\mathcal{H}}
        |\hat{X}_{ct}-X_{ct}|.
    \end{aligned}
\end{equation}
Here, $\hat{X}_{ct}$ and $X_{ct}$ denote the imputed value and its corresponding ground truth, respectively. Restricting both the reconstruction loss and the evaluation metrics to $\mathcal{H}$ ensures that performance reflects the recovery of artificially masked entries and is not diluted by the substantially more numerous observed entries.

\input{Appendix/Tables/Efficiency}

\paragraph{Training Details.}
We train GLAIM by minimizing the mean squared error (MSE) over the artificially masked positions $\mathcal{H}$. For each input window, entries selected by the artificial mask are replaced with zeros, while the corresponding binary mask is provided to the model to distinguish masked entries from observed values. The optimization objective is computed exclusively on $\mathcal{H}$, so observed entries do not contribute to the reconstruction loss.

GLAIM is implemented in PyTorch~\cite{PyTroch} and trained with a batch size of 16. We use Adam~\cite{Adam} with $(\beta_1,\beta_2,\epsilon)=(0.9,0.999,10^{-8})$, zero weight decay, and a fixed learning rate. The learning rate is $3\times10^{-4}$ for Electricity, Weather, and PEMS03, and $10^{-3}$ for the other datasets. 
Training proceeds for at most 300 epochs, and the model is evaluated on the validation set after each epoch. Early stopping is applied with a patience of 30 epochs according to the validation MSE computed over artificially masked positions, and the checkpoint achieving the lowest validation error is used for final testing. All experiments are run on NVIDIA RTX 4090 GPUs.

\subsection{Baseline Implementation Details}
The complete comparison contains two categories of baselines: general-purpose time series models, including TimeDART~\cite{TimeDART}, TimesNet~\cite{TimesNet}, iTransformer~\cite{YLiu2024c}, ModernTCN~\cite{ModernTCN}, PatchTST~\cite{PatchTST}, DLinear~\cite{DLinear}, and FreTS~\cite{FreTS}; and specialized imputation methods, including ImputeFormer~\cite{TNie2024b}, SAITS~\cite{WDu2023h}, and PSW-I~\cite{PSWI}. 
All methods use the same chronological data splits, artificial-missingness settings, and masked-position evaluation protocol. General-purpose baselines use the shared training settings described above, whereas specialized imputation methods retain their original optimization procedures and method-specific training settings.

For general-purpose models that contain both a hidden representation dimension $d_{\mathrm{model}}$ and a feed-forward dimension $d_{\mathrm{ff}}$, we set
\begin{equation}
    d_{\mathrm{model}} = 128
    \quad\text{and}\quad
    d_{\mathrm{ff}} = 256,
\end{equation}
which matches the corresponding configuration used by GLAIM. 
Applying the same representation and feed-forward dimensions reduces the influence of model capacity and enables a more direct comparison between different architectural designs. 
Parameters specific to each architecture are otherwise inherited from its official implementation.

For every baseline, we directly adopt the dataset-specific configuration released by the official implementation whenever available, while applying the explicitly standardized parameters above where applicable. 
If the target dataset is unsupported, we select the official benchmark configuration whose number of variables is closest to that of the target dataset. 
The number of variables is used as the primary matching criterion because it directly affects the dimensionality and computational cost of cross-variable modeling. 
This rule provides a consistent and reproducible procedure for adapting both general-purpose and specialized baselines to unsupported datasets without method-specific manual tuning.

Except for the explicitly standardized parameters, we preserve the original architectural choices and implementation details of each baseline as closely as possible. All models are trained and evaluated within the same experimental framework, while method-specific components follow their official implementations. This protocol is intended to balance comparability and fidelity: shared experimental factors are controlled across all methods, whereas the defining design choices of each baseline are retained.
Except for the explicitly standardized parameters, the original architectural and optimization choices of each baseline are preserved as closely as possible. Thus, data exposure and evaluation are controlled across methods, while method-specific designs and training protocols are retained.

\section{Efficiency Analysis}
For a controlled efficiency comparison, methods parameterized by $d_{\mathrm{model}}$ and $d_{\mathrm{ff}}$ are configured with 64 and 128, respectively. This compact common setting limits the extent to which representation width dominates resource consumption, allowing the comparison to more directly reflect architecture-induced overhead, while the main experiments retain the larger 128/256 configuration for performance evaluation.

To evaluate the computational overhead of GLAIM, Table~\ref{tab:efficiency} compares the parameter count, GPU memory consumption, computational cost, and per-iteration training time of different models on ETTh1 and Weather.
Across both datasets, GLAIM maintains a moderate model size, requiring only $0.479$M parameters on ETTh1 and $0.543$M parameters on Weather. Compared with ImputeFormer, GLAIM reduces the parameter count by $26.3\%$ and $28.3\%$, the memory consumption by $29.7\%$ and $23.0\%$, and the computational cost by $52.3\%$ and $51.6\%$ on ETTh1 and Weather, respectively. GLAIM is also substantially more compact than TimesNet, using more than $94\%$ fewer parameters and $89.0\%$--$96.4\%$ fewer GFLOPs.

In terms of per-iteration training time, GLAIM requires $21.351$ ms/iter on ETTh1 and $24.870$ ms/iter on Weather. Its training time on ETTh1 is comparable to that of TimesNet ($21.770$ ms/iter), whereas it is higher than those of TimesNet and ImputeFormer on Weather. GLAIM also has a higher per-iteration time than lightweight architectures such as DLinear, FreTS, iTransformer, and PatchTST. This runtime overhead is consistent with the additional operations required to construct stable global dependencies and perform sample-conditioned dependency refinement, which are not captured by the parameter count alone.

Therefore, GLAIM is not the most lightweight model in terms of either memory consumption or iteration time. Nevertheless, its parameter count remains below $0.55$M on both datasets, and its computational cost is considerably lower than that of intensive alternatives such as ImputeFormer and TimesNet. Taken together with its strong imputation accuracy, these results indicate that GLAIM provides a reasonable accuracy--efficiency trade-off: it incurs additional runtime overhead over simple baselines to explicitly model cross-variable dependencies, while avoiding the substantially larger parameter and computation costs of several competitive methods.

\section{Hyperparameter Sensitivity}
\begin{figure}[t]
    \centering
    \includegraphics[width=1.0\columnwidth]{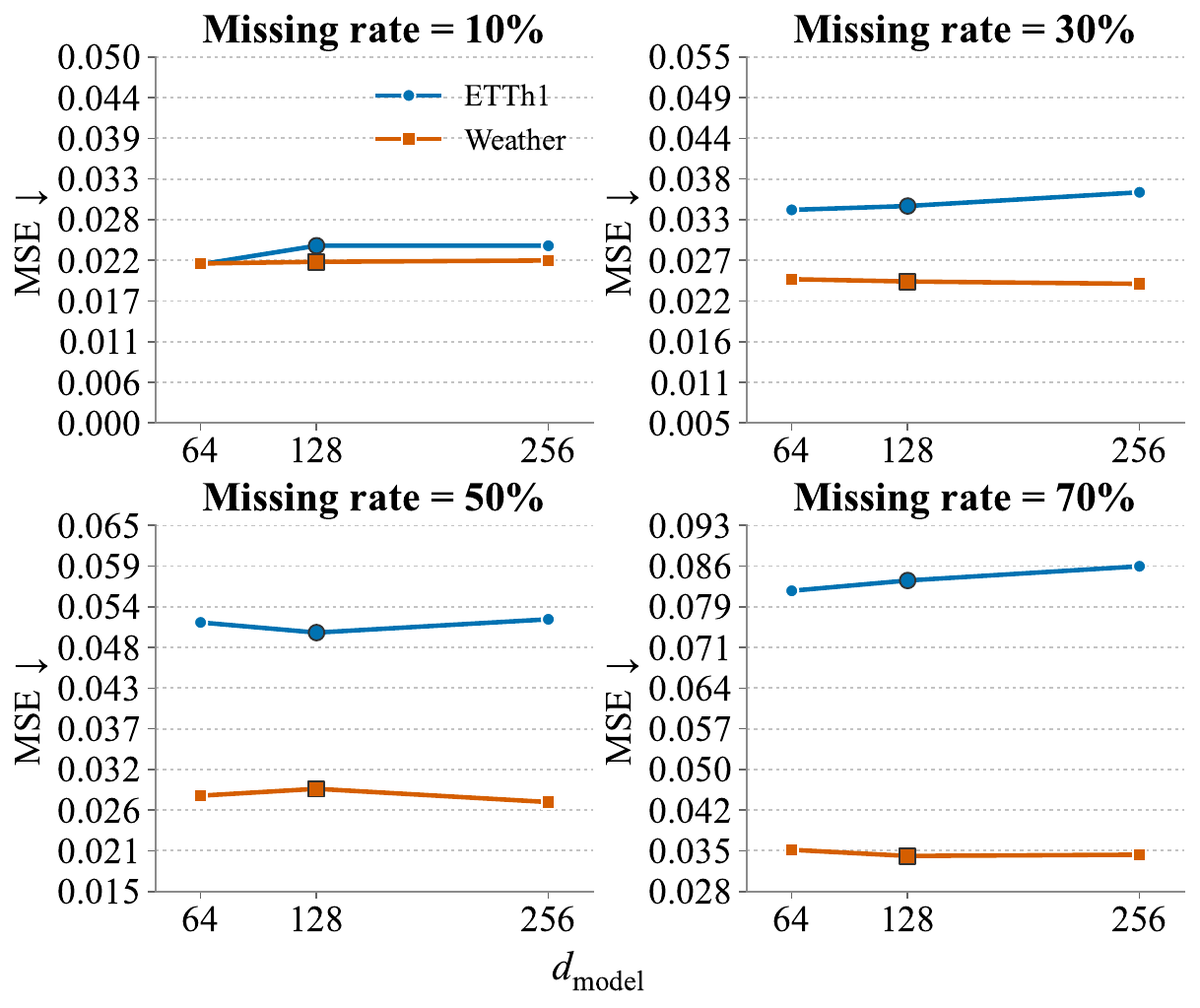}
    \caption{Sensitivity to the hidden dimension $D$ on ETTh1 and Weather under random missingness. MSE is reported at missing rates of $0.1$, $0.3$, $0.5$, and $0.7$; lower values indicate better performance.}
    \label{fig:sensitivity_d_model}
\end{figure}

\begin{figure}[t]
    \centering
    \includegraphics[width=1.0\columnwidth]{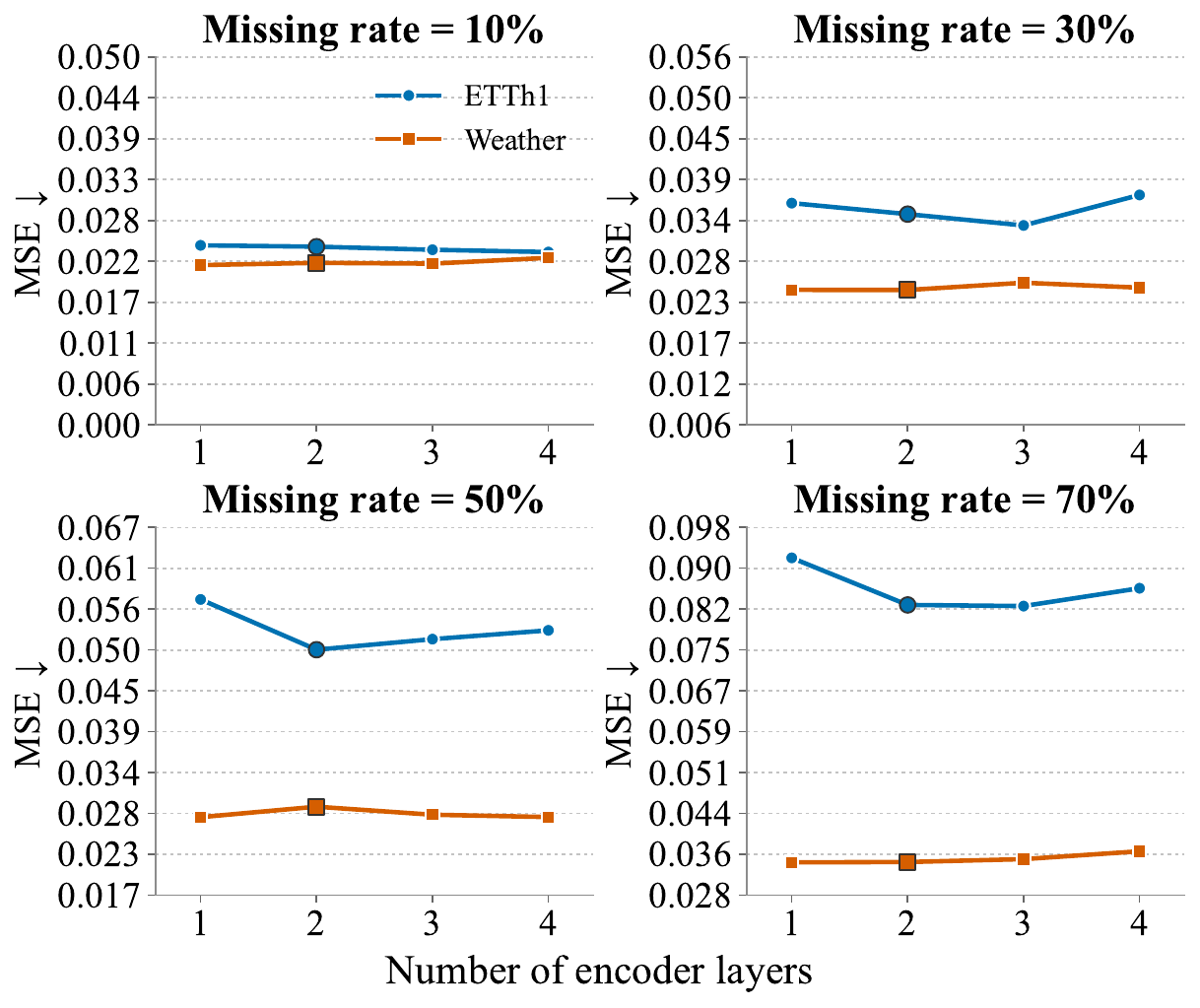}
    \caption{Sensitivity to the number of GloLoc layers $L$ on ETTh1 and Weather under random missingness. A moderate depth of two or three layers consistently provides the best overall performance.}
    \label{fig:sensitivity_e_layers}
\end{figure}

\begin{figure}[t]
    \centering
    \includegraphics[width=1.0\columnwidth]{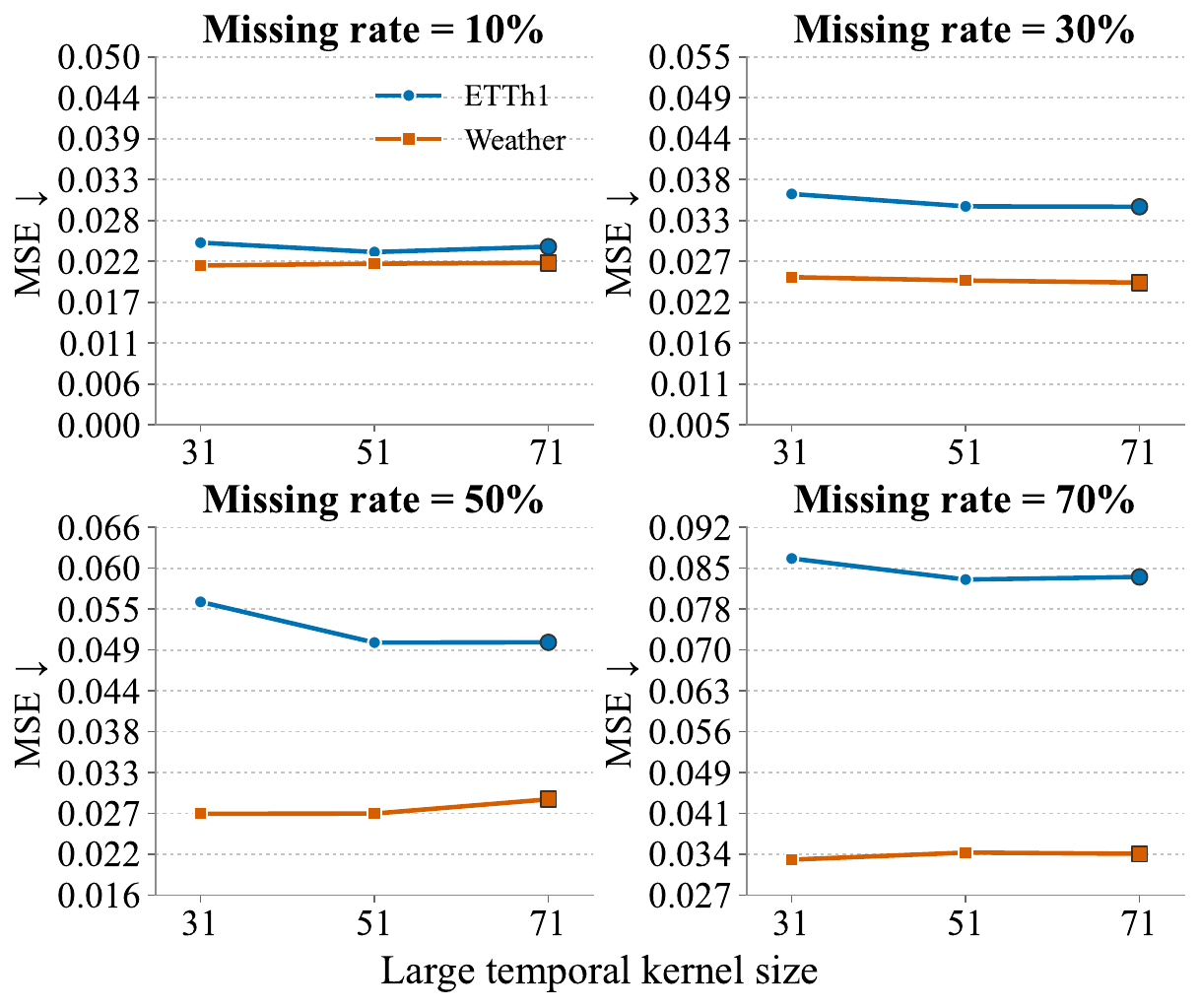}
    \caption{Sensitivity to the large temporal kernel size $k_{\mathrm{L}}$ in StaGlo on ETTh1 and Weather under random missingness. Larger kernels generally improve performance on ETTh1, while Weather is comparatively insensitive to the kernel size.}
    \label{fig:sensitivity_large_kernel}
\end{figure}

We investigate the sensitivity of GLAIM to three key architectural hyperparameters: the hidden dimension $D$, the number of GloLoc layers $L$, and the large temporal kernel size $k_{\mathrm{L}}$ in StaGlo. Experiments are conducted on ETTh1 and Weather under random missingness rates of $0.1$, $0.3$, $0.5$, and $0.7$. We vary one hyperparameter at a time while keeping the remaining settings fixed. The reference configuration used in the main experiments is $D=128$, $L=2$, and $k_{\mathrm{L}}=71$.

\paragraph{Hidden Dimension.}
Figure~\ref{fig:sensitivity_d_model} reports the results for $D\in\{64,128,256\}$. The performance remains stable across the tested dimensions, with average MSE values of $0.0374$, $0.0378$, and $0.0383$, respectively, over the two datasets and four missing rates. Increasing the hidden dimension does not consistently reduce the error: a smaller dimension performs better in several ETTh1 settings, whereas the relative ordering varies slightly on Weather. The difference between the best and worst average MSE is only $2.6\%$, indicating that GLAIM is not sensitive to the representation width within this range. We use $D=128$ as a moderately sized default that provides consistently competitive performance.

\paragraph{Encoder Depth.}
Figure~\ref{fig:sensitivity_e_layers} compares $L\in\{1,2,3,4\}$. Using only one GloLoc layer produces an average MSE of $0.0397$ and is particularly less effective on ETTh1 at high missing rates. Increasing the depth to two or three layers improves the average MSE to $0.0378$ and $0.0377$, respectively. In contrast, four layers increase the average MSE to $0.0389$, showing that the benefit of stacking additional global--local dependency encoding layers saturates at a moderate depth. Since two and three layers achieve nearly identical performance, we adopt $L=2$ to retain a compact architecture.

\paragraph{Large Temporal Kernel.}
Figure~\ref{fig:sensitivity_large_kernel} evaluates $k_{\mathrm{L}}\in\{31,51,71\}$. The corresponding average MSE values are $0.0388$, $0.0374$, and $0.0378$. On ETTh1, the larger kernels of $51$ and $71$ consistently outperform $31$ at medium and high missing rates, indicating that a sufficiently broad temporal receptive field is beneficial when local observations become sparse. Weather exhibits smaller and non-monotonic differences, suggesting that its temporal patterns require less long-range aggregation. Overall, both $51$ and $71$ provide robust results, and the default value $k_{\mathrm{L}}=71$ remains competitive across all evaluated settings without dataset-specific tuning.

Overall, the results demonstrate that GLAIM is robust to reasonable variations in its principal architectural hyperparameters. The best-to-worst differences in average MSE are $2.6\%$, $5.3\%$, and $3.7\%$ for the hidden dimension, encoder depth, and large temporal kernel size, respectively. The selected configuration therefore provides stable performance rather than relying on narrowly tuned hyperparameter values.

\section{Additional Dependency Visualizations}
\begin{figure}[t]
    \centering
    \includegraphics[width=0.85\columnwidth]{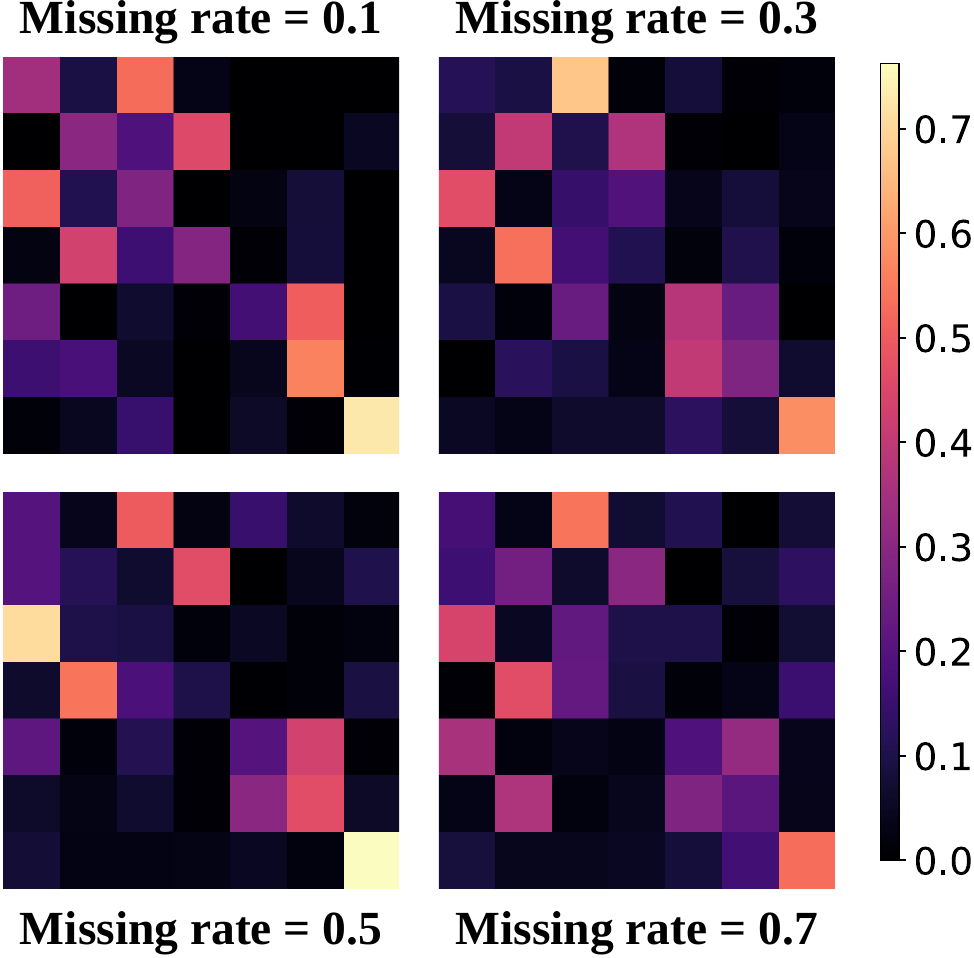}
    \caption{Global dependency matrices from the first StaGlo layer on ETTh1 under block missingness. Despite changes in the missing rate, the matrices retain a shared cross-variable dependency structure.}
    \label{fig:dependency_1}
\end{figure}

\begin{figure}[t]
    \centering
    \includegraphics[width=0.85\columnwidth]{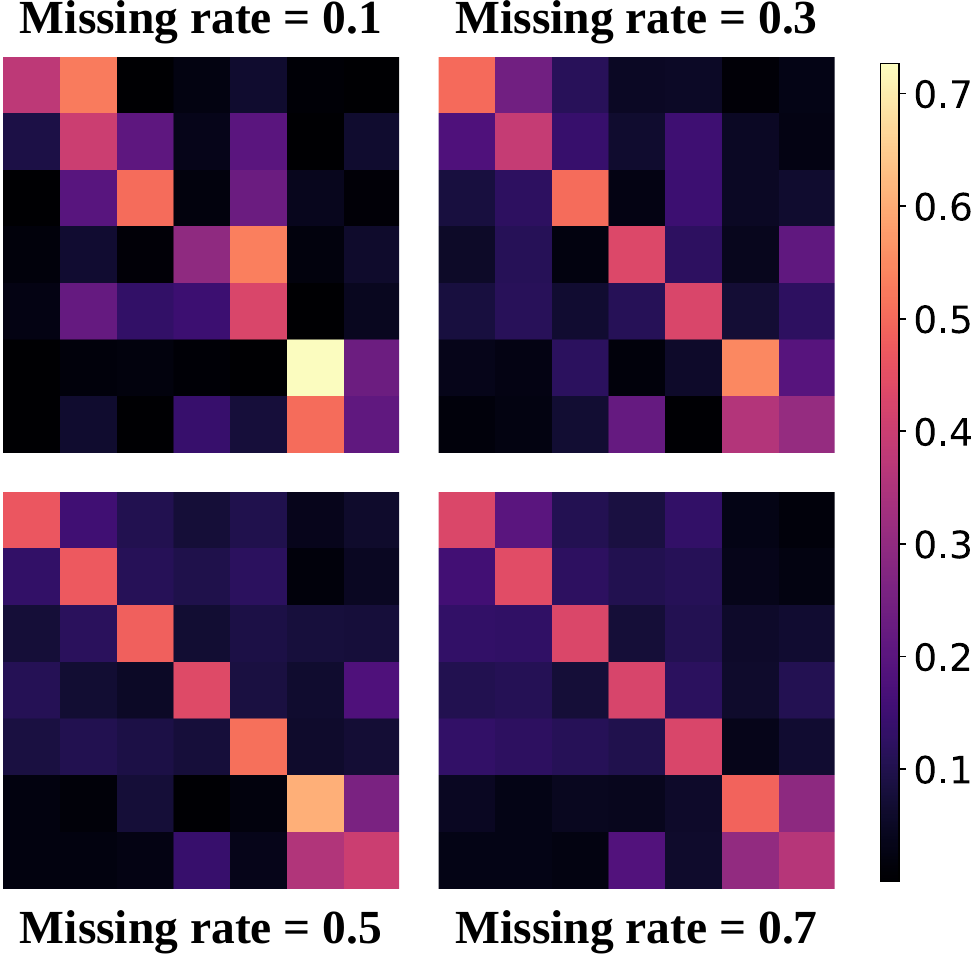}
    \caption{Global dependency matrices from the first StaGlo layer on Illness under random missingness. The learned inter-variable structures remain visually consistent across missing rates.}
    \label{fig:dependency_2}
\end{figure}

We provide two dependency visualizations covering different missingness patterns and datasets. Specifically, we independently train one GLAIM model at each missing rate in $\{0.1,0.3,0.5,0.7\}$ and visualize the aggregated dependency matrices produced by the first StaGlo layer. Figure~\ref{fig:dependency_1} considers block missingness on ETTh1, whereas Figure~\ref{fig:dependency_2} considers random missingness on Illness. The corresponding pairwise Spearman correlations and Jensen--Shannon divergences are reported in Tables~\ref{tab:dependency_1} and~\ref{tab:dependency_2}, respectively.

\input{Appendix/Tables/dependency_1}

\input{Appendix/Tables/dependency_2}

As shown in Figure~\ref{fig:dependency_1}, the dominant dependency patterns remain recognizable across block missing rates. Quantitatively, the off-diagonal pairwise comparisons yield Spearman correlations ranging from $0.564$ to $0.735$ and Jensen--Shannon divergences ranging from $0.072$ to $0.143$. Even for the largest shift, from a missing rate of $0.1$ to $0.7$, the matrices retain a Spearman correlation of $0.564$ with a Jensen--Shannon divergence of $0.143$. These results suggest that the global dependency backbone remains identifiable under contiguous missing segments, although block missingness introduces greater variation than the ETTh1 random-missing case.

Figure~\ref{fig:dependency_2} shows similarly stable dependency structures on Illness. Across all off-diagonal comparisons, the Spearman correlations range from $0.737$ to $0.901$, while the Jensen--Shannon divergences range from $0.014$ to $0.103$. In particular, the matrices learned at missing rates of $0.5$ and $0.7$ achieve a Spearman correlation of $0.901$ and a Jensen--Shannon divergence of only $0.014$. The comparison between the extreme missing rates of $0.1$ and $0.7$ also remains consistent, with a Spearman correlation of $0.775$ and a Jensen--Shannon divergence of $0.102$.

Taken together, these additional cases show that StaGlo recovers persistent global dependency structures across datasets and missingness patterns. The Illness results demonstrate strong stability under random missingness, while the ETTh1 block-missing results indicate that the learned dependency backbone remains robust even when observations are removed in contiguous segments.

\section{Full Results}
This section provides the complete comparison with all baselines used in our experiments, including TimeDART~\cite{TimeDART}, ModernTCN~\cite{ModernTCN}, iTransformer~\cite{YLiu2024c}, TimesNet~\cite{TimesNet}, PatchTST~\cite{PatchTST}, DLinear~\cite{DLinear}, FreTS~\cite{FreTS}, ImputeFormer~\cite{TNie2024b}, SAITS~\cite{WDu2023h}, and PSW-I~\cite{PSWI} where applicable. The tables report both MSE and MAE, with lower values indicating better performance.

\paragraph{Random Missing.}
\input{Appendix/Tables/random_full}
As shown in Table~\ref{tab:random_full}, the complete random-missing results demonstrate a consistent advantage across datasets and missing rates. Averaged over the four missing rates, GLAIM achieves the lowest MSE and MAE on all nine datasets, with overall averages of $0.030$ and $0.089$, reducing MSE and MAE by $50.0\%$ and $32.1\%$, respectively, compared with ModernTCN, the strongest baseline on average. GLAIM also obtains the best or tied-best result in $68$ of the $72$ dataset--rate--metric comparisons; the only exceptions occur on ETTh1 at missing rates of $0.1$ and $0.3$, where SAITS is marginally better.

\paragraph{Block Missing.}
\input{Appendix/Tables/multi_block_full}
As shown in Table~\ref{tab:block_full}, the complete block-missing results exhibit a similarly consistent advantage. Averaged over the four missing rates, GLAIM ranks first in all $18$ dataset--metric comparisons, with overall MSE and MAE values of $0.050$ and $0.113$, corresponding to reductions of $55.0\%$ and $33.9\%$, respectively, over ModernTCN. At individual missing rates, GLAIM achieves the best or tied-best result in $70$ of the $72$ comparisons, with the only exceptions occurring on ETTh1 at a missing rate of $0.1$, where SAITS obtains slightly lower MSE and MAE.

\paragraph{Missing-Rate Shift.}
We provide complete robustness results under both random and block missing-rate shifts. All trainable models are trained at a missing rate of $0.3$ and evaluated at rates of $0.1$, $0.3$, $0.5$, and $0.7$, where $0.1$, $0.5$, and $0.7$ are unseen during training. PSW-I is excluded because it is training-free.

\input{Appendix/Tables/random_robust_full}
As shown in Table~\ref{tab:random_robust}, GLAIM achieves the best average MSE and MAE at every test missing rate under random missingness. Compared with the strongest baseline at each rate, it reduces MSE by $54.2\%$--$64.2\%$ and MAE by $33.6\%$--$43.2\%$. At the highest unseen rate of $0.7$, the reductions remain $64.2\%$ and $41.0\%$, respectively.
\input{Appendix/Tables/multi_block_robust_full}
As shown in Table~\ref{tab:block_robust}, GLAIM also achieves the best average MSE and MAE at every test rate under block missingness. Its improvements over the strongest baseline range from $50.6\%$ to $60.3\%$ in MSE and from $30.5\%$ to $41.3\%$ in MAE. Even at the unseen rate of $0.7$, GLAIM reduces MSE and MAE by $50.6\%$ and $30.5\%$, respectively.

Overall, the complete results confirm that GLAIM's advantages are preserved across the full collection of baselines, datasets, missingness patterns, and missing rates. The improvements are particularly pronounced as the missingness condition becomes more challenging, supporting the effectiveness of combining stable global dependency modeling with sample-conditioned local refinement.

%% file: Appendix/Tables/dataset_statistics.tex
\begin{table*}[t]
\centering
\caption{Dataset descriptions.}
\begin{tabular}{lcccccc}
\toprule
\textbf{Dataset} & \textbf{Variables} & \textbf{Train} & \textbf{Valid} & \textbf{Test} & \textbf{Frequency} & \textbf{Complete time range} \\
\midrule
ETTh1,ETTh2 & 7 & 8,640 & 2,880 & 5,760 & Hourly & 2016-07-01 00:00 – 2018-06-26 19:00 \\
ETTm1,ETTm2 & 7 & 34,560 & 11,520 & 23,040 & 15min & 2016-07-01 00:00 – 2018-06-26 19:45 \\
Electricity (ECL) & 321 & 18,412 & 2,632 & 5,260 & Hourly & 2016-07-01 02:00 – 2019-07-02 01:00 \\
Exchange & 8 & 5,311 & 760 & 1,517 & Daily & 1990-01-01 – 2010-10-10 \\
ILI & 7 & 676 & 97 & 193 & Weekly & 2002-01-01 – 2020-06-30 \\
Weather & 21 & 36,887 & 5,270 & 10,539 & 10min & 2020-01-01 00:10 – 2021-01-01 00:00 \\
PEMS03 & 358 & 18,345 & 2,622 & 5,241 & 5min & 2018-09-01 00:00 – 2018-11-30 23:55 \\
\bottomrule
\end{tabular}
\label{tab:dataset_statistics}
\end{table*}

%% file: Appendix/Tables/Efficiency.tex
\begin{table*}[t]
\centering
\caption{Efficiency comparison on ETTh1 and Weather in terms of parameter count, GPU memory consumption, computational cost, and per-iteration training speed. Lower values indicate lower resource requirements and faster training.}
\begin{tabular}{c|l|c|c|c|c}
\toprule
Dataset & Model & Params (M) & Memory (MiB) & GFLOPs & Train Speed (ms/iter) \\
\midrule
\multirow{10}{*}{ETTh1} & \textbf{GLAIM (Ours)} & 0.479 & 786.7 & 5.381 & 21.351 \\
 & TimeDART & 0.395 & 264.4 & 0.904 & 6.325 \\
 & ModernTCN & 1.697 & 184.0 & 1.219 & 6.451 \\
 & iTransformer & 0.080 & 21.5 & 0.028 & 7.136 \\
 & TimesNet & 9.376 & 284.9 & 149.353 & 21.770 \\
 & PatchTST & 0.142 & 46.9 & 0.204 & 7.373 \\
 & DLinear & 0.019 & 17.0 & 0.004 & 2.361 \\
 & FreTS & 3.237 & 140.0 & 0.710 & 3.527 \\
 & ImputeFormer & 0.650 & 1119.0 & 11.274 & 18.160 \\
 & SAITS & 0.333 & 115.2 & 1.621 & 10.875 \\
\midrule
\multirow{10}{*}{Weather} & \textbf{GLAIM (Ours)} & 0.543 & 1924.6 & 16.381 &  24.870 \\
 & TimeDART & 0.395 & 496.1 & 2.713 & 13.505 \\
 & ModernTCN & 2.406 & 432.2 & 3.889 & 7.215 \\
 & iTransformer & 0.080 & 27.5 & 0.067 & 6.916 \\
 & TimesNet & 9.380 & 300.7 & 149.364 & 19.771 \\
 & PatchTST & 0.142 & 104.1 & 0.611 & 7.955 \\
 & DLinear & 0.019 & 18.0 & 0.012 & 2.471 \\
 & FreTS & 3.237 & 290.5 & 2.130 & 4.016 \\
 & ImputeFormer & 0.757 & 2500.2 & 33.823 & 18.502 \\
 & SAITS & 0.341 & 117.0 & 1.644 & 9.984 \\
\bottomrule
\end{tabular}
\label{tab:efficiency}
\end{table*}

%% file: Appendix/Tables/dependency_1.tex
\begin{table*}[t]
\centering
\caption{Pairwise Spearman correlations and Jensen--Shannon divergences between the first-layer StaGlo dependency matrices learned on ETTh1 under block missingness at different missing rates.}
\begin{tabular}{ccccccccc}
\toprule
\multirow{2}{*}{Missing Ratio (MR)} & \multicolumn{4}{c}{Spearman correlations} & \multicolumn{4}{c}{Jensen--Shannon divergences} \\
\cmidrule(lr){2-5}
\cmidrule(lr){6-9}
 & MR=0.1 & MR=0.3 & MR=0.5 & MR=0.7 & MR=0.1 & MR=0.3 & MR=0.5 & MR=0.7 \\
\midrule
0.1 & 1.000 & 0.728 & 0.671 & 0.564 & 0.000 & 0.126 & 0.110 & 0.143 \\
0.3 & -- & 1.000 & 0.685 & 0.679 & -- & 0.000 & 0.080 & 0.080 \\
0.5 & -- & -- & 1.000 & 0.735 & -- & -- & 0.000 & 0.072 \\
0.7 & -- & -- & -- & 1.000 & -- & -- & -- & 0.000 \\
\bottomrule
\end{tabular}
\label{tab:dependency_1}
\end{table*}

%% file: Appendix/Tables/dependency_2.tex
\begin{table*}[t]
\centering
\caption{Pairwise Spearman correlations and Jensen--Shannon divergences between the first-layer StaGlo dependency matrices learned on Illness under random missingness at different missing rates.}
\begin{tabular}{ccccccccc}
\toprule
\multirow{2}{*}{Missing Ratio (MR)} & \multicolumn{4}{c}{Spearman correlations} & \multicolumn{4}{c}{Jensen--Shannon divergences} \\
\cmidrule(lr){2-5}
\cmidrule(lr){6-9}
 & MR=0.1 & MR=0.3 & MR=0.5 & MR=0.7 & MR=0.1 & MR=0.3 & MR=0.5 & MR=0.7 \\
\midrule
0.1 & 1.000 & 0.737 & 0.779 & 0.775 & 0.000 & 0.096 & 0.103 & 0.102 \\
0.3 & -- & 1.000 & 0.885 & 0.837 & -- & 0.000 & 0.020 & 0.024 \\
0.5 & -- & -- & 1.000 & 0.901 & -- & -- & 0.000 & 0.014 \\
0.7 & -- & -- & -- & 1.000 & -- & -- & -- & 0.000 \\
\bottomrule
\end{tabular}
\label{tab:dependency_2}
\end{table*}

%% file: Appendix/Tables/random_full.tex
\begin{table*}[t]
\centering
\scriptsize
\setlength{\tabcolsep}{2pt}
\caption{Full results under random missingness at missing rates of $0.1$, $0.3$, $0.5$, and $0.7$ across datasets.}
\resizebox{\textwidth}{!}{%
\begin{tabular}{lr|rr|rr|rr|rr|rr|rr|rr|rr|rr|rr|rr}
\toprule
\multicolumn{2}{c|}{Models} & \multicolumn{2}{c|}{\textbf{GLAIM (Ours)}} & \multicolumn{2}{c|}{TimeDART} & \multicolumn{2}{c|}{ModernTCN} & \multicolumn{2}{c|}{iTransformer} & \multicolumn{2}{c|}{TimesNet} & \multicolumn{2}{c|}{PatchTST} & \multicolumn{2}{c|}{DLinear} & \multicolumn{2}{c|}{FreTS} & \multicolumn{2}{c|}{ImputeFormer} & \multicolumn{2}{c|}{SAITS} & \multicolumn{2}{c}{PSW-I} \\
\multicolumn{2}{c|}{Metirc} & MSE & MAE & MSE & MAE & MSE & MAE & MSE & MAE & MSE & MAE & MSE & MAE & MSE & MAE & MSE & MAE & MSE & MAE & MSE & MAE & MSE & MAE \\
\midrule
\multirow{5}{*}{\rotatebox{90}{ETTh1}} & 0.1 & \underline{\textcolor{blue}{0.024}} & \underline{\textcolor{blue}{0.106}} & 0.091 & 0.198 & 0.031 & 0.122 & 0.081 & 0.201 & 0.044 & 0.145 & 0.074 & 0.179 & 0.111 & 0.232 & 0.101 & 0.219 & 0.050 & 0.139 & \textcolor{red}{\textbf{0.022}} & \textcolor{red}{\textbf{0.101}} & 0.067 & 0.168 \\
 & 0.3 & \underline{\textcolor{blue}{0.035}} & \underline{\textcolor{blue}{0.122}} & 0.116 & 0.223 & 0.046 & 0.145 & 0.118 & 0.243 & 0.074 & 0.185 & 0.089 & 0.196 & 0.169 & 0.287 & 0.140 & 0.259 & 0.060 & 0.155 & \textcolor{red}{\textbf{0.033}} & \textcolor{red}{\textbf{0.118}} & 0.084 & 0.186 \\
 & 0.5 & \textcolor{red}{\textbf{0.050}} & \textcolor{red}{\textbf{0.144}} & 0.145 & 0.249 & 0.068 & 0.175 & 0.157 & 0.279 & 0.101 & 0.216 & 0.115 & 0.221 & 0.234 & 0.334 & 0.186 & 0.296 & 0.097 & 0.200 & \underline{\textcolor{blue}{0.067}} & \underline{\textcolor{blue}{0.166}} & 0.105 & 0.208 \\
 & 0.7 & \textcolor{red}{\textbf{0.083}} & \textcolor{red}{\textbf{0.184}} & 0.238 & 0.315 & \underline{\textcolor{blue}{0.107}} & \underline{\textcolor{blue}{0.217}} & 0.192 & 0.304 & 0.151 & 0.260 & 0.165 & 0.264 & 0.325 & 0.390 & 0.259 & 0.347 & 0.193 & 0.271 & 0.143 & 0.267 & 0.209 & 0.271 \\
 & Avg & \textcolor{red}{\textbf{0.048}} & \textcolor{red}{\textbf{0.139}} & 0.148 & 0.246 & \underline{\textcolor{blue}{0.063}} & 0.165 & 0.137 & 0.257 & 0.093 & 0.201 & 0.111 & 0.215 & 0.210 & 0.311 & 0.171 & 0.280 & 0.100 & 0.191 & 0.066 & \underline{\textcolor{blue}{0.163}} & 0.116 & 0.208 \\
\midrule
\multirow{5}{*}{\rotatebox{90}{ETTh2}} & 0.1 & \textcolor{red}{\textbf{0.027}} & \textcolor{red}{\textbf{0.093}} & 0.060 & 0.154 & 0.036 & \underline{\textcolor{blue}{0.119}} & 0.074 & 0.184 & \underline{\textcolor{blue}{0.036}} & 0.122 & 0.054 & 0.144 & 0.117 & 0.231 & 0.083 & 0.193 & 0.068 & 0.158 & 0.060 & 0.159 & 0.046 & 0.130 \\
 & 0.3 & \textcolor{red}{\textbf{0.031}} & \textcolor{red}{\textbf{0.104}} & 0.067 & 0.167 & \underline{\textcolor{blue}{0.041}} & \underline{\textcolor{blue}{0.130}} & 0.112 & 0.230 & 0.049 & 0.148 & 0.060 & 0.152 & 0.169 & 0.280 & 0.110 & 0.223 & 0.071 & 0.156 & 0.087 & 0.192 & 0.054 & 0.142 \\
 & 0.5 & \textcolor{red}{\textbf{0.039}} & \textcolor{red}{\textbf{0.119}} & 0.078 & 0.182 & \underline{\textcolor{blue}{0.049}} & \underline{\textcolor{blue}{0.144}} & 0.142 & 0.261 & 0.055 & 0.154 & 0.067 & 0.163 & 0.223 & 0.324 & 0.129 & 0.245 & 0.121 & 0.220 & 0.138 & 0.229 & 0.061 & 0.150 \\
 & 0.7 & \textcolor{red}{\textbf{0.054}} & \textcolor{red}{\textbf{0.143}} & 0.097 & 0.204 & \underline{\textcolor{blue}{0.064}} & \underline{\textcolor{blue}{0.166}} & 0.166 & 0.283 & 0.073 & 0.177 & 0.084 & 0.185 & 0.295 & 0.375 & 0.147 & 0.263 & 0.159 & 0.265 & 0.207 & 0.315 & 0.075 & 0.170 \\
 & Avg & \textcolor{red}{\textbf{0.038}} & \textcolor{red}{\textbf{0.115}} & 0.075 & 0.177 & \underline{\textcolor{blue}{0.048}} & \underline{\textcolor{blue}{0.140}} & 0.124 & 0.240 & 0.053 & 0.150 & 0.066 & 0.161 & 0.201 & 0.302 & 0.117 & 0.231 & 0.105 & 0.200 & 0.123 & 0.224 & 0.059 & 0.148 \\
\midrule
\multirow{5}{*}{\rotatebox{90}{ETTm1}} & 0.1 & \textcolor{red}{\textbf{0.013}} & \textcolor{red}{\textbf{0.074}} & 0.035 & 0.116 & 0.015 & 0.081 & 0.040 & 0.135 & 0.019 & 0.093 & 0.036 & 0.120 & 0.057 & 0.164 & 0.048 & 0.148 & 0.015 & 0.082 & \underline{\textcolor{blue}{0.013}} & \underline{\textcolor{blue}{0.075}} & 0.035 & 0.117 \\
 & 0.3 & \textcolor{red}{\textbf{0.016}} & \textcolor{red}{\textbf{0.080}} & 0.040 & 0.127 & 0.020 & 0.091 & 0.057 & 0.164 & 0.027 & 0.110 & 0.041 & 0.129 & 0.088 & 0.206 & 0.064 & 0.172 & 0.021 & 0.092 & \underline{\textcolor{blue}{0.017}} & \underline{\textcolor{blue}{0.089}} & 0.035 & 0.119 \\
 & 0.5 & \textcolor{red}{\textbf{0.021}} & \textcolor{red}{\textbf{0.092}} & 0.050 & 0.142 & 0.027 & 0.106 & 0.097 & 0.215 & 0.035 & 0.124 & 0.047 & 0.138 & 0.129 & 0.248 & 0.081 & 0.193 & 0.027 & 0.106 & \underline{\textcolor{blue}{0.023}} & \underline{\textcolor{blue}{0.099}} & 0.047 & 0.131 \\
 & 0.7 & \textcolor{red}{\textbf{0.032}} & \textcolor{red}{\textbf{0.114}} & 0.071 & 0.169 & 0.042 & 0.131 & 0.095 & 0.211 & 0.054 & 0.150 & 0.063 & 0.159 & 0.203 & 0.308 & 0.109 & 0.221 & \underline{\textcolor{blue}{0.038}} & \underline{\textcolor{blue}{0.124}} & 0.041 & 0.133 & 0.060 & 0.154 \\
 & Avg & \textcolor{red}{\textbf{0.021}} & \textcolor{red}{\textbf{0.090}} & 0.049 & 0.138 & 0.026 & 0.102 & 0.072 & 0.181 & 0.034 & 0.119 & 0.047 & 0.136 & 0.119 & 0.232 & 0.075 & 0.183 & 0.025 & 0.101 & \underline{\textcolor{blue}{0.024}} & \underline{\textcolor{blue}{0.099}} & 0.044 & 0.130 \\
\midrule
\multirow{5}{*}{\rotatebox{90}{ETTm2}} & 0.1 & \textcolor{red}{\textbf{0.012}} & \textcolor{red}{\textbf{0.054}} & 0.025 & 0.092 & \underline{\textcolor{blue}{0.017}} & \underline{\textcolor{blue}{0.075}} & 0.037 & 0.123 & 0.018 & 0.078 & 0.024 & 0.088 & 0.074 & 0.180 & 0.044 & 0.133 & 0.033 & 0.092 & 0.023 & 0.098 & 0.023 & 0.084 \\
 & 0.3 & \textcolor{red}{\textbf{0.015}} & \textcolor{red}{\textbf{0.062}} & 0.031 & 0.105 & \underline{\textcolor{blue}{0.019}} & \underline{\textcolor{blue}{0.081}} & 0.050 & 0.146 & 0.021 & 0.088 & 0.028 & 0.097 & 0.108 & 0.220 & 0.047 & 0.139 & 0.030 & 0.092 & 0.028 & 0.106 & 0.030 & 0.093 \\
 & 0.5 & \textcolor{red}{\textbf{0.018}} & \textcolor{red}{\textbf{0.072}} & 0.037 & 0.118 & \underline{\textcolor{blue}{0.023}} & \underline{\textcolor{blue}{0.091}} & 0.065 & 0.170 & 0.026 & 0.098 & 0.032 & 0.105 & 0.146 & 0.258 & 0.055 & 0.152 & 0.034 & 0.102 & 0.050 & 0.152 & 0.031 & 0.100 \\
 & 0.7 & \textcolor{red}{\textbf{0.026}} & \textcolor{red}{\textbf{0.090}} & 0.046 & 0.131 & \underline{\textcolor{blue}{0.030}} & \underline{\textcolor{blue}{0.105}} & 0.093 & 0.204 & 0.032 & 0.112 & 0.040 & 0.118 & 0.204 & 0.308 & 0.070 & 0.173 & 0.046 & 0.134 & 0.072 & 0.184 & 0.038 & 0.113 \\
 & Avg & \textcolor{red}{\textbf{0.018}} & \textcolor{red}{\textbf{0.070}} & 0.035 & 0.111 & \underline{\textcolor{blue}{0.022}} & \underline{\textcolor{blue}{0.088}} & 0.061 & 0.161 & 0.024 & 0.094 & 0.031 & 0.102 & 0.133 & 0.241 & 0.054 & 0.149 & 0.036 & 0.105 & 0.043 & 0.135 & 0.031 & 0.097 \\
\midrule
\multirow{5}{*}{\rotatebox{90}{Weather}} & 0.1 & \textcolor{red}{\textbf{0.022}} & \textcolor{red}{\textbf{0.031}} & 0.028 & 0.047 & \underline{\textcolor{blue}{0.024}} & 0.040 & 0.033 & 0.076 & 0.026 & 0.047 & 0.028 & 0.048 & 0.037 & 0.086 & 0.033 & 0.075 & 0.036 & 0.075 & 0.037 & 0.090 & 0.025 & \underline{\textcolor{blue}{0.040}} \\
 & 0.3 & \textcolor{red}{\textbf{0.024}} & \textcolor{red}{\textbf{0.035}} & 0.030 & 0.047 & \underline{\textcolor{blue}{0.025}} & \underline{\textcolor{blue}{0.040}} & 0.039 & 0.084 & 0.029 & 0.055 & 0.031 & 0.051 & 0.051 & 0.115 & 0.043 & 0.095 & 0.036 & 0.065 & 0.034 & 0.073 & 0.027 & 0.042 \\
 & 0.5 & \textcolor{red}{\textbf{0.029}} & \textcolor{red}{\textbf{0.042}} & 0.033 & 0.049 & \underline{\textcolor{blue}{0.029}} & \underline{\textcolor{blue}{0.046}} & 0.042 & 0.084 & 0.036 & 0.066 & 0.033 & 0.050 & 0.066 & 0.137 & 0.052 & 0.109 & 0.039 & 0.073 & 0.044 & 0.097 & 0.036 & 0.047 \\
 & 0.7 & \textcolor{red}{\textbf{0.034}} & \textcolor{red}{\textbf{0.047}} & 0.042 & 0.057 & \underline{\textcolor{blue}{0.037}} & 0.057 & 0.057 & 0.102 & 0.041 & 0.070 & 0.042 & 0.060 & 0.090 & 0.168 & 0.065 & 0.126 & 0.047 & 0.083 & 0.050 & 0.107 & 0.048 & \underline{\textcolor{blue}{0.055}} \\
 & Avg & \textcolor{red}{\textbf{0.027}} & \textcolor{red}{\textbf{0.039}} & 0.033 & 0.050 & \underline{\textcolor{blue}{0.029}} & \underline{\textcolor{blue}{0.046}} & 0.043 & 0.087 & 0.033 & 0.060 & 0.034 & 0.052 & 0.061 & 0.127 & 0.048 & 0.101 & 0.040 & 0.074 & 0.041 & 0.092 & 0.034 & 0.046 \\
\midrule
\multirow{5}{*}{\rotatebox{90}{PEMS03}} & 0.1 & \textcolor{red}{\textbf{0.013}} & \textcolor{red}{\textbf{0.072}} & 0.034 & 0.129 & 0.023 & 0.100 & 0.031 & 0.123 & 0.032 & 0.120 & 0.036 & 0.134 & 0.045 & 0.154 & 0.036 & 0.133 & \underline{\textcolor{blue}{0.019}} & \underline{\textcolor{blue}{0.087}} & 0.051 & 0.129 & 0.038 & 0.137 \\
 & 0.3 & \textcolor{red}{\textbf{0.015}} & \textcolor{red}{\textbf{0.078}} & 0.034 & 0.126 & 0.028 & 0.113 & 0.038 & 0.133 & 0.036 & 0.130 & 0.038 & 0.138 & 0.063 & 0.185 & 0.040 & 0.139 & \underline{\textcolor{blue}{0.021}} & \underline{\textcolor{blue}{0.092}} & 0.061 & 0.144 & 0.038 & 0.136 \\
 & 0.5 & \textcolor{red}{\textbf{0.018}} & \textcolor{red}{\textbf{0.085}} & 0.037 & 0.132 & 0.033 & 0.124 & 0.046 & 0.143 & 0.043 & 0.143 & 0.042 & 0.146 & 0.085 & 0.216 & 0.045 & 0.147 & \underline{\textcolor{blue}{0.033}} & \underline{\textcolor{blue}{0.120}} & 0.066 & 0.154 & 0.042 & 0.143 \\
 & 0.7 & \textcolor{red}{\textbf{0.023}} & \textcolor{red}{\textbf{0.097}} & 0.044 & 0.142 & \underline{\textcolor{blue}{0.041}} & \underline{\textcolor{blue}{0.139}} & 0.052 & 0.156 & 0.054 & 0.163 & 0.050 & 0.157 & 0.124 & 0.263 & 0.056 & 0.164 & 0.051 & 0.152 & 0.074 & 0.167 & 0.047 & 0.150 \\
 & Avg & \textcolor{red}{\textbf{0.017}} & \textcolor{red}{\textbf{0.083}} & 0.037 & 0.132 & 0.031 & 0.119 & 0.042 & 0.139 & 0.041 & 0.139 & 0.042 & 0.144 & 0.079 & 0.204 & 0.045 & 0.146 & \underline{\textcolor{blue}{0.031}} & \underline{\textcolor{blue}{0.113}} & 0.063 & 0.149 & 0.041 & 0.141 \\
\midrule
\multirow{5}{*}{\rotatebox{90}{Exchange}} & 0.1 & \textcolor{red}{\textbf{0.002}} & \textcolor{red}{\textbf{0.018}} & 0.003 & 0.030 & 0.003 & 0.027 & 0.006 & 0.048 & \underline{\textcolor{blue}{0.002}} & 0.025 & 0.002 & 0.026 & 0.024 & 0.108 & 0.013 & 0.080 & 0.017 & 0.072 & 0.008 & 0.061 & 0.003 & \underline{\textcolor{blue}{0.024}} \\
 & 0.3 & \textcolor{red}{\textbf{0.002}} & \textcolor{red}{\textbf{0.019}} & 0.003 & 0.035 & 0.003 & 0.028 & 0.010 & 0.066 & 0.003 & 0.032 & 0.003 & 0.028 & 0.046 & 0.148 & 0.012 & 0.075 & 0.022 & 0.079 & 0.022 & 0.106 & \underline{\textcolor{blue}{0.002}} & \underline{\textcolor{blue}{0.026}} \\
 & 0.5 & \textcolor{red}{\textbf{0.002}} & \textcolor{red}{\textbf{0.023}} & 0.004 & 0.036 & 0.003 & 0.033 & 0.033 & 0.124 & 0.004 & 0.037 & 0.003 & 0.033 & 0.072 & 0.186 & 0.011 & 0.072 & 0.027 & 0.099 & 0.111 & 0.248 & \underline{\textcolor{blue}{0.003}} & \underline{\textcolor{blue}{0.030}} \\
 & 0.7 & \textcolor{red}{\textbf{0.003}} & \textcolor{red}{\textbf{0.030}} & 0.005 & 0.043 & 0.005 & 0.042 & 0.021 & 0.099 & 0.006 & 0.047 & \underline{\textcolor{blue}{0.004}} & 0.039 & 0.120 & 0.240 & 0.013 & 0.076 & 0.028 & 0.104 & 0.130 & 0.302 & 0.005 & \underline{\textcolor{blue}{0.035}} \\
 & Avg & \textcolor{red}{\textbf{0.002}} & \textcolor{red}{\textbf{0.023}} & 0.004 & 0.036 & 0.003 & 0.032 & 0.017 & 0.085 & 0.004 & 0.035 & 0.003 & 0.031 & 0.065 & 0.170 & 0.012 & 0.075 & 0.023 & 0.088 & 0.068 & 0.179 & \underline{\textcolor{blue}{0.003}} & \underline{\textcolor{blue}{0.029}} \\
\midrule
\multirow{5}{*}{\rotatebox{90}{Illness}} & 0.1 & \textcolor{red}{\textbf{0.016}} & \textcolor{red}{\textbf{0.067}} & 0.115 & 0.206 & 0.108 & 0.209 & 0.227 & 0.307 & 0.132 & 0.227 & 0.098 & 0.192 & 0.254 & 0.300 & 0.329 & 0.369 & 0.515 & 0.406 & 0.244 & 0.258 & \underline{\textcolor{blue}{0.095}} & \underline{\textcolor{blue}{0.128}} \\
 & 0.3 & \textcolor{red}{\textbf{0.026}} & \textcolor{red}{\textbf{0.085}} & 0.175 & 0.255 & 0.178 & 0.257 & 0.681 & 0.575 & 0.164 & 0.252 & \underline{\textcolor{blue}{0.153}} & 0.239 & 0.471 & 0.410 & 0.453 & 0.415 & 0.699 & 0.479 & 0.341 & 0.315 & 0.191 & \underline{\textcolor{blue}{0.205}} \\
 & 0.5 & \textcolor{red}{\textbf{0.052}} & \textcolor{red}{\textbf{0.120}} & 0.282 & 0.310 & 0.263 & 0.314 & 0.937 & 0.655 & \underline{\textcolor{blue}{0.226}} & 0.299 & 0.265 & 0.299 & 0.710 & 0.521 & 0.706 & 0.491 & 1.121 & 0.607 & 0.633 & 0.465 & 0.267 & \underline{\textcolor{blue}{0.245}} \\
 & 0.7 & \textcolor{red}{\textbf{0.108}} & \textcolor{red}{\textbf{0.170}} & 0.467 & 0.398 & 0.396 & 0.392 & 1.056 & 0.702 & \underline{\textcolor{blue}{0.312}} & \underline{\textcolor{blue}{0.345}} & 0.469 & 0.405 & 1.021 & 0.619 & 1.189 & 0.649 & 1.409 & 0.718 & 0.740 & 0.523 & 0.830 & 0.420 \\
 & Avg & \textcolor{red}{\textbf{0.051}} & \textcolor{red}{\textbf{0.110}} & 0.260 & 0.292 & 0.236 & 0.293 & 0.726 & 0.560 & \underline{\textcolor{blue}{0.209}} & 0.281 & 0.246 & 0.284 & 0.614 & 0.462 & 0.669 & 0.481 & 0.936 & 0.552 & 0.490 & 0.390 & 0.346 & \underline{\textcolor{blue}{0.250}} \\
\midrule
\multirow{5}{*}{\rotatebox{90}{Electricity}} & 0.1 & \textcolor{red}{\textbf{0.033}} & \textcolor{red}{\textbf{0.115}} & 0.042 & 0.131 & 0.055 & 0.163 & 0.052 & 0.153 & 0.091 & 0.206 & 0.047 & 0.147 & 0.078 & 0.198 & 0.060 & 0.164 & \underline{\textcolor{blue}{0.038}} & \underline{\textcolor{blue}{0.127}} & 0.126 & 0.248 & 0.045 & 0.139 \\
 & 0.3 & \textcolor{red}{\textbf{0.038}} & \textcolor{red}{\textbf{0.124}} & 0.056 & 0.158 & 0.074 & 0.188 & 0.075 & 0.188 & 0.097 & 0.213 & 0.059 & 0.165 & 0.125 & 0.256 & 0.088 & 0.204 & \underline{\textcolor{blue}{0.040}} & \underline{\textcolor{blue}{0.128}} & 0.163 & 0.278 & 0.054 & 0.153 \\
 & 0.5 & \textcolor{red}{\textbf{0.047}} & \textcolor{red}{\textbf{0.138}} & 0.073 & 0.180 & 0.084 & 0.199 & 0.095 & 0.213 & 0.105 & 0.224 & 0.076 & 0.191 & 0.174 & 0.305 & 0.114 & 0.235 & \underline{\textcolor{blue}{0.059}} & \underline{\textcolor{blue}{0.158}} & 0.187 & 0.298 & 0.067 & 0.173 \\
 & 0.7 & \textcolor{red}{\textbf{0.061}} & \textcolor{red}{\textbf{0.157}} & 0.103 & 0.210 & 0.102 & 0.219 & 0.121 & 0.239 & 0.122 & 0.245 & 0.110 & 0.224 & 0.247 & 0.365 & 0.153 & 0.274 & \underline{\textcolor{blue}{0.075}} & \underline{\textcolor{blue}{0.180}} & 0.206 & 0.308 & 0.094 & 0.206 \\
 & Avg & \textcolor{red}{\textbf{0.045}} & \textcolor{red}{\textbf{0.133}} & 0.068 & 0.170 & 0.079 & 0.192 & 0.086 & 0.198 & 0.103 & 0.222 & 0.073 & 0.182 & 0.156 & 0.281 & 0.104 & 0.219 & \underline{\textcolor{blue}{0.053}} & \underline{\textcolor{blue}{0.148}} & 0.171 & 0.283 & 0.065 & 0.168 \\
\bottomrule
\end{tabular}
}
\label{tab:random_full}
\end{table*}

%% file: Appendix/Tables/multi_block_full.tex
\begin{table*}[t]
\centering
\scriptsize
\setlength{\tabcolsep}{2pt}
\caption{Full results under block missingness at missing rates of $0.1$, $0.3$, $0.5$, and $0.7$ across datasets.}
\resizebox{\textwidth}{!}{%
\begin{tabular}{lr|rr|rr|rr|rr|rr|rr|rr|rr|rr|rr|rr}
\toprule
\multicolumn{2}{c|}{Models} & \multicolumn{2}{c|}{\textbf{GLAIM (Ours)}} & \multicolumn{2}{c|}{TimeDART} & \multicolumn{2}{c|}{ModernTCN} & \multicolumn{2}{c|}{iTransformer} & \multicolumn{2}{c|}{TimesNet} & \multicolumn{2}{c|}{PatchTST} & \multicolumn{2}{c|}{DLinear} & \multicolumn{2}{c|}{FreTS} & \multicolumn{2}{c|}{ImputeFormer} & \multicolumn{2}{c|}{SAITS} & \multicolumn{2}{c}{PSW-I} \\
\multicolumn{2}{c|}{Metric} & MSE & MAE & MSE & MAE & MSE & MAE & MSE & MAE & MSE & MAE & MSE & MAE & MSE & MAE & MSE & MAE & MSE & MAE & MSE & MAE & MSE & MAE \\
\midrule
\multirow{5}{*}{\rotatebox{90}{ETTh1}} & 0.1 & \underline{\textcolor{blue}{0.030}} & \underline{\textcolor{blue}{0.116}} & 0.168 & 0.268 & 0.047 & 0.150 & 0.139 & 0.259 & 0.078 & 0.191 & 0.123 & 0.231 & 0.226 & 0.329 & 0.187 & 0.296 & 0.084 & 0.177 & \textcolor{red}{\textbf{0.027}} & \textcolor{red}{\textbf{0.110}} & 0.352 & 0.331 \\
 & 0.3 & \textcolor{red}{\textbf{0.054}} & \textcolor{red}{\textbf{0.150}} & 0.237 & 0.315 & 0.090 & 0.200 & 0.217 & 0.319 & 0.132 & 0.241 & 0.188 & 0.281 & 0.317 & 0.381 & 0.277 & 0.354 & 0.151 & 0.234 & \underline{\textcolor{blue}{0.069}} & \underline{\textcolor{blue}{0.159}} & 0.877 & 0.536 \\
 & 0.5 & \textcolor{red}{\textbf{0.089}} & \textcolor{red}{\textbf{0.189}} & 0.265 & 0.332 & 0.130 & 0.237 & 0.258 & 0.347 & 0.200 & 0.291 & 0.206 & 0.292 & 0.379 & 0.416 & 0.313 & 0.376 & 0.220 & 0.299 & \underline{\textcolor{blue}{0.128}} & \underline{\textcolor{blue}{0.217}} & 0.746 & 0.513 \\
 & 0.7 & \textcolor{red}{\textbf{0.133}} & \textcolor{red}{\textbf{0.228}} & 0.336 & 0.377 & \underline{\textcolor{blue}{0.184}} & \underline{\textcolor{blue}{0.279}} & 0.299 & 0.369 & 0.291 & 0.348 & 0.256 & 0.326 & 0.460 & 0.457 & 0.366 & 0.403 & 0.315 & 0.353 & 0.210 & 0.288 & 1.259 & 0.682 \\
 & Avg & \textcolor{red}{\textbf{0.076}} & \textcolor{red}{\textbf{0.171}} & 0.252 & 0.323 & 0.113 & 0.217 & 0.228 & 0.323 & 0.175 & 0.268 & 0.193 & 0.283 & 0.345 & 0.395 & 0.286 & 0.357 & 0.192 & 0.266 & \underline{\textcolor{blue}{0.109}} & \underline{\textcolor{blue}{0.194}} & 0.809 & 0.516 \\
\midrule
\multirow{5}{*}{\rotatebox{90}{ETTh2}} & 0.1 & \textcolor{red}{\textbf{0.029}} & \textcolor{red}{\textbf{0.100}} & 0.081 & 0.187 & \underline{\textcolor{blue}{0.046}} & \underline{\textcolor{blue}{0.141}} & 0.089 & 0.204 & 0.048 & 0.147 & 0.072 & 0.172 & 0.168 & 0.286 & 0.107 & 0.221 & 0.090 & 0.188 & 0.064 & 0.173 & 0.289 & 0.343 \\
 & 0.3 & \textcolor{red}{\textbf{0.043}} & \textcolor{red}{\textbf{0.128}} & 0.107 & 0.218 & \underline{\textcolor{blue}{0.061}} & \underline{\textcolor{blue}{0.166}} & 0.143 & 0.264 & 0.071 & 0.179 & 0.092 & 0.198 & 0.232 & 0.336 & 0.142 & 0.259 & 0.127 & 0.223 & 0.181 & 0.298 & 0.290 & 0.342 \\
 & 0.5 & \textcolor{red}{\textbf{0.053}} & \textcolor{red}{\textbf{0.145}} & 0.121 & 0.232 & \underline{\textcolor{blue}{0.073}} & \underline{\textcolor{blue}{0.182}} & 0.156 & 0.277 & 0.089 & 0.200 & 0.101 & 0.209 & 0.249 & 0.352 & 0.150 & 0.266 & 0.209 & 0.299 & 0.240 & 0.341 & 0.265 & 0.329 \\
 & 0.7 & \textcolor{red}{\textbf{0.073}} & \textcolor{red}{\textbf{0.170}} & 0.136 & 0.246 & \underline{\textcolor{blue}{0.090}} & \underline{\textcolor{blue}{0.202}} & 0.177 & 0.292 & 0.108 & 0.219 & 0.116 & 0.225 & 0.256 & 0.357 & 0.163 & 0.277 & 0.272 & 0.353 & 0.310 & 0.384 & 0.329 & 0.371 \\
 & Avg & \textcolor{red}{\textbf{0.049}} & \textcolor{red}{\textbf{0.136}} & 0.111 & 0.221 & \underline{\textcolor{blue}{0.067}} & \underline{\textcolor{blue}{0.173}} & 0.141 & 0.259 & 0.079 & 0.186 & 0.095 & 0.201 & 0.226 & 0.333 & 0.141 & 0.256 & 0.175 & 0.266 & 0.199 & 0.299 & 0.293 & 0.346 \\
\midrule
\multirow{5}{*}{\rotatebox{90}{ETTm1}} & 0.1 & \textcolor{red}{\textbf{0.014}} & \textcolor{red}{\textbf{0.079}} & 0.053 & 0.147 & 0.019 & 0.093 & 0.065 & 0.175 & 0.027 & 0.112 & 0.051 & 0.145 & 0.108 & 0.229 & 0.080 & 0.190 & 0.024 & 0.101 & \underline{\textcolor{blue}{0.015}} & \underline{\textcolor{blue}{0.080}} & 0.732 & 0.480 \\
 & 0.3 & \textcolor{red}{\textbf{0.023}} & \textcolor{red}{\textbf{0.096}} & 0.088 & 0.192 & 0.034 & 0.123 & 0.119 & 0.235 & 0.051 & 0.151 & 0.075 & 0.175 & 0.240 & 0.333 & 0.141 & 0.249 & 0.036 & 0.123 & \underline{\textcolor{blue}{0.028}} & \underline{\textcolor{blue}{0.108}} & 1.135 & 0.610 \\
 & 0.5 & \textcolor{red}{\textbf{0.034}} & \textcolor{red}{\textbf{0.116}} & 0.106 & 0.209 & 0.052 & 0.149 & 0.159 & 0.271 & 0.076 & 0.178 & 0.092 & 0.193 & 0.310 & 0.374 & 0.175 & 0.275 & 0.051 & 0.146 & \underline{\textcolor{blue}{0.040}} & \underline{\textcolor{blue}{0.131}} & 1.093 & 0.614 \\
 & 0.7 & \textcolor{red}{\textbf{0.051}} & \textcolor{red}{\textbf{0.142}} & 0.137 & 0.231 & 0.078 & 0.179 & 0.152 & 0.260 & 0.132 & 0.225 & 0.125 & 0.220 & 0.357 & 0.400 & 0.210 & 0.297 & 0.076 & 0.179 & \underline{\textcolor{blue}{0.072}} & \underline{\textcolor{blue}{0.171}} & 1.026 & 0.612 \\
 & Avg & \textcolor{red}{\textbf{0.031}} & \textcolor{red}{\textbf{0.108}} & 0.096 & 0.195 & 0.046 & 0.136 & 0.124 & 0.235 & 0.072 & 0.166 & 0.086 & 0.184 & 0.254 & 0.334 & 0.151 & 0.253 & 0.047 & 0.137 & \underline{\textcolor{blue}{0.039}} & \underline{\textcolor{blue}{0.123}} & 0.996 & 0.579 \\
\midrule
\multirow{5}{*}{\rotatebox{90}{ETTm2}} & 0.1 & \textcolor{red}{\textbf{0.015}} & \textcolor{red}{\textbf{0.063}} & 0.042 & 0.125 & \underline{\textcolor{blue}{0.023}} & \underline{\textcolor{blue}{0.089}} & 0.048 & 0.142 & 0.024 & 0.092 & 0.037 & 0.114 & 0.094 & 0.209 & 0.057 & 0.154 & 0.038 & 0.100 & 0.028 & 0.110 & 0.172 & 0.239 \\
 & 0.3 & \textcolor{red}{\textbf{0.021}} & \textcolor{red}{\textbf{0.076}} & 0.055 & 0.147 & \underline{\textcolor{blue}{0.029}} & \underline{\textcolor{blue}{0.105}} & 0.070 & 0.175 & 0.033 & 0.116 & 0.049 & 0.134 & 0.141 & 0.262 & 0.068 & 0.173 & 0.055 & 0.135 & 0.039 & 0.131 & 0.207 & 0.285 \\
 & 0.5 & \textcolor{red}{\textbf{0.026}} & \textcolor{red}{\textbf{0.091}} & 0.065 & 0.164 & \underline{\textcolor{blue}{0.035}} & \underline{\textcolor{blue}{0.117}} & 0.081 & 0.191 & 0.038 & 0.124 & 0.054 & 0.141 & 0.148 & 0.272 & 0.077 & 0.186 & 0.064 & 0.147 & 0.091 & 0.200 & 0.252 & 0.303 \\
 & 0.7 & \textcolor{red}{\textbf{0.035}} & \textcolor{red}{\textbf{0.108}} & 0.067 & 0.163 & \underline{\textcolor{blue}{0.043}} & \underline{\textcolor{blue}{0.130}} & 0.098 & 0.213 & 0.050 & 0.140 & 0.061 & 0.150 & 0.151 & 0.274 & 0.084 & 0.194 & 0.075 & 0.175 & 0.116 & 0.231 & 0.227 & 0.296 \\
 & Avg & \textcolor{red}{\textbf{0.024}} & \textcolor{red}{\textbf{0.085}} & 0.057 & 0.150 & \underline{\textcolor{blue}{0.032}} & \underline{\textcolor{blue}{0.110}} & 0.074 & 0.180 & 0.036 & 0.118 & 0.050 & 0.135 & 0.133 & 0.254 & 0.071 & 0.177 & 0.058 & 0.139 & 0.069 & 0.168 & 0.215 & 0.281 \\
\midrule
\multirow{5}{*}{\rotatebox{90}{Weather}} & 0.1 & \textcolor{red}{\textbf{0.028}} & \textcolor{red}{\textbf{0.038}} & 0.037 & 0.052 & \underline{\textcolor{blue}{0.031}} & \underline{\textcolor{blue}{0.049}} & 0.049 & 0.090 & 0.036 & 0.057 & 0.042 & 0.064 & 0.057 & 0.111 & 0.052 & 0.098 & 0.049 & 0.092 & 0.045 & 0.093 & 0.145 & 0.150 \\
 & 0.3 & \textcolor{red}{\textbf{0.035}} & \textcolor{red}{\textbf{0.048}} & 0.049 & 0.066 & \underline{\textcolor{blue}{0.038}} & \underline{\textcolor{blue}{0.058}} & 0.065 & 0.110 & 0.045 & 0.075 & 0.052 & 0.072 & 0.085 & 0.154 & 0.064 & 0.112 & 0.050 & 0.081 & 0.059 & 0.117 & 0.133 & 0.167 \\
 & 0.5 & \textcolor{red}{\textbf{0.040}} & \textcolor{red}{\textbf{0.053}} & 0.053 & 0.071 & \underline{\textcolor{blue}{0.043}} & \underline{\textcolor{blue}{0.068}} & 0.068 & 0.113 & 0.055 & 0.084 & 0.053 & 0.072 & 0.099 & 0.172 & 0.076 & 0.134 & 0.051 & 0.084 & 0.062 & 0.118 & 0.202 & 0.201 \\
 & 0.7 & \textcolor{red}{\textbf{0.043}} & \textcolor{red}{\textbf{0.054}} & 0.059 & 0.078 & \underline{\textcolor{blue}{0.049}} & \underline{\textcolor{blue}{0.077}} & 0.076 & 0.120 & 0.062 & 0.094 & 0.060 & 0.079 & 0.109 & 0.185 & 0.075 & 0.123 & 0.054 & 0.084 & 0.067 & 0.125 & 0.184 & 0.211 \\
 & Avg & \textcolor{red}{\textbf{0.036}} & \textcolor{red}{\textbf{0.049}} & 0.049 & 0.067 & \underline{\textcolor{blue}{0.041}} & \underline{\textcolor{blue}{0.063}} & 0.065 & 0.108 & 0.049 & 0.078 & 0.052 & 0.072 & 0.087 & 0.156 & 0.067 & 0.117 & 0.051 & 0.085 & 0.058 & 0.113 & 0.166 & 0.182 \\
\midrule
\multirow{5}{*}{\rotatebox{90}{PEMS03}} & 0.1 & \textcolor{red}{\textbf{0.014}} & \textcolor{red}{\textbf{0.074}} & 0.038 & 0.133 & 0.026 & 0.106 & 0.036 & 0.129 & 0.034 & 0.125 & 0.040 & 0.139 & 0.061 & 0.177 & 0.043 & 0.142 & \underline{\textcolor{blue}{0.024}} & \underline{\textcolor{blue}{0.095}} & 0.054 & 0.133 & 0.147 & 0.238 \\
 & 0.3 & \textcolor{red}{\textbf{0.017}} & \textcolor{red}{\textbf{0.081}} & 0.046 & 0.145 & 0.037 & 0.128 & 0.049 & 0.147 & 0.051 & 0.157 & 0.048 & 0.150 & 0.109 & 0.242 & 0.057 & 0.162 & \underline{\textcolor{blue}{0.025}} & \underline{\textcolor{blue}{0.098}} & 0.063 & 0.144 & 0.319 & 0.357 \\
 & 0.5 & \textcolor{red}{\textbf{0.020}} & \textcolor{red}{\textbf{0.089}} & 0.050 & 0.149 & 0.042 & 0.139 & 0.056 & 0.154 & 0.068 & 0.185 & 0.055 & 0.159 & 0.141 & 0.276 & 0.066 & 0.175 & \underline{\textcolor{blue}{0.032}} & \underline{\textcolor{blue}{0.111}} & 0.070 & 0.160 & 0.488 & 0.455 \\
 & 0.7 & \textcolor{red}{\textbf{0.025}} & \textcolor{red}{\textbf{0.101}} & 0.059 & 0.161 & 0.050 & 0.153 & 0.061 & 0.165 & 0.087 & 0.210 & 0.065 & 0.173 & 0.170 & 0.306 & 0.078 & 0.190 & \underline{\textcolor{blue}{0.039}} & \underline{\textcolor{blue}{0.123}} & 0.075 & 0.168 & 0.641 & 0.540 \\
 & Avg & \textcolor{red}{\textbf{0.019}} & \textcolor{red}{\textbf{0.086}} & 0.048 & 0.147 & 0.039 & 0.132 & 0.051 & 0.149 & 0.060 & 0.169 & 0.052 & 0.155 & 0.120 & 0.250 & 0.061 & 0.167 & \underline{\textcolor{blue}{0.030}} & \underline{\textcolor{blue}{0.107}} & 0.065 & 0.151 & 0.399 & 0.398 \\
\midrule
\multirow{5}{*}{\rotatebox{90}{Exchange}} & 0.1 & \textcolor{red}{\textbf{0.003}} & \textcolor{red}{\textbf{0.025}} & 0.005 & 0.043 & 0.005 & 0.040 & 0.008 & 0.057 & 0.005 & 0.038 & \underline{\textcolor{blue}{0.004}} & \underline{\textcolor{blue}{0.035}} & 0.028 & 0.117 & 0.009 & 0.062 & 0.018 & 0.082 & 0.027 & 0.120 & 0.007 & 0.060 \\
 & 0.3 & \textcolor{red}{\textbf{0.004}} & \textcolor{red}{\textbf{0.033}} & 0.006 & 0.050 & 0.006 & 0.050 & 0.012 & 0.073 & 0.008 & 0.054 & \underline{\textcolor{blue}{0.006}} & \underline{\textcolor{blue}{0.046}} & 0.039 & 0.136 & 0.012 & 0.075 & 0.030 & 0.102 & 0.128 & 0.301 & 0.019 & 0.087 \\
 & 0.5 & \textcolor{red}{\textbf{0.005}} & \textcolor{red}{\textbf{0.038}} & 0.008 & 0.055 & 0.008 & 0.056 & 0.019 & 0.093 & 0.009 & 0.061 & \underline{\textcolor{blue}{0.007}} & \underline{\textcolor{blue}{0.051}} & 0.043 & 0.146 & 0.013 & 0.079 & 0.038 & 0.118 & 0.195 & 0.372 & 0.024 & 0.103 \\
 & 0.7 & \textcolor{red}{\textbf{0.006}} & \textcolor{red}{\textbf{0.044}} & 0.009 & 0.060 & 0.009 & 0.061 & 0.016 & 0.087 & 0.011 & 0.067 & \underline{\textcolor{blue}{0.008}} & \underline{\textcolor{blue}{0.056}} & 0.042 & 0.146 & 0.016 & 0.086 & 0.046 & 0.138 & 0.206 & 0.403 & 0.037 & 0.117 \\
 & Avg & \textcolor{red}{\textbf{0.004}} & \textcolor{red}{\textbf{0.035}} & 0.007 & 0.052 & 0.007 & 0.052 & 0.014 & 0.077 & 0.008 & 0.055 & \underline{\textcolor{blue}{0.006}} & \underline{\textcolor{blue}{0.047}} & 0.038 & 0.136 & 0.013 & 0.075 & 0.033 & 0.110 & 0.139 & 0.299 & 0.022 & 0.091 \\
\midrule
\multirow{5}{*}{\rotatebox{90}{Illness}} & 0.1 & \textcolor{red}{\textbf{0.046}} & \textcolor{red}{\textbf{0.113}} & 0.352 & 0.344 & \underline{\textcolor{blue}{0.201}} & 0.279 & 0.449 & 0.432 & 0.206 & \underline{\textcolor{blue}{0.276}} & 0.295 & 0.312 & 0.630 & 0.499 & 0.844 & 0.543 & 0.740 & 0.493 & 0.285 & 0.289 & 0.335 & 0.381 \\
 & 0.3 & \textcolor{red}{\textbf{0.093}} & \textcolor{red}{\textbf{0.160}} & 0.656 & 0.492 & 0.419 & 0.405 & 0.876 & 0.617 & \underline{\textcolor{blue}{0.404}} & \underline{\textcolor{blue}{0.392}} & 0.650 & 0.483 & 1.223 & 0.699 & 2.028 & 0.880 & 1.065 & 0.620 & 0.635 & 0.498 & 2.748 & 0.892 \\
 & 0.5 & \textcolor{red}{\textbf{0.156}} & \textcolor{red}{\textbf{0.207}} & 0.731 & 0.488 & 0.629 & 0.487 & 1.157 & 0.689 & \underline{\textcolor{blue}{0.589}} & \underline{\textcolor{blue}{0.459}} & 0.791 & 0.517 & 1.448 & 0.750 & 2.244 & 0.920 & 1.423 & 0.697 & 0.645 & 0.475 & 1.229 & 0.703 \\
 & 0.7 & \textcolor{red}{\textbf{0.266}} & \textcolor{red}{\textbf{0.275}} & 0.898 & 0.556 & 0.940 & 0.594 & 1.447 & 0.767 & \underline{\textcolor{blue}{0.840}} & \underline{\textcolor{blue}{0.549}} & 0.960 & 0.577 & 1.768 & 0.824 & 2.524 & 0.981 & 1.875 & 0.787 & 1.140 & 0.633 & 3.161 & 1.102 \\
 & Avg & \textcolor{red}{\textbf{0.140}} & \textcolor{red}{\textbf{0.189}} & 0.659 & 0.470 & 0.547 & 0.441 & 0.982 & 0.626 & \underline{\textcolor{blue}{0.510}} & \underline{\textcolor{blue}{0.419}} & 0.674 & 0.472 & 1.267 & 0.693 & 1.910 & 0.831 & 1.276 & 0.649 & 0.676 & 0.474 & 1.868 & 0.770 \\
\midrule
\multirow{5}{*}{\rotatebox{90}{Electricity}} & 0.1 & \textcolor{red}{\textbf{0.046}} & \underline{\textcolor{blue}{0.137}} & 0.062 & 0.158 & 0.084 & 0.201 & 0.075 & 0.183 & 0.101 & 0.216 & 0.064 & 0.171 & 0.125 & 0.246 & 0.088 & 0.199 & \underline{\textcolor{blue}{0.047}} & \textcolor{red}{\textbf{0.137}} & 0.181 & 0.292 & 0.141 & 0.243 \\
 & 0.3 & \textcolor{red}{\textbf{0.060}} & \textcolor{red}{\textbf{0.155}} & 0.089 & 0.192 & 0.104 & 0.220 & 0.104 & 0.217 & 0.115 & 0.236 & 0.086 & 0.195 & 0.197 & 0.313 & 0.129 & 0.241 & \underline{\textcolor{blue}{0.061}} & \underline{\textcolor{blue}{0.157}} & 0.194 & 0.305 & 0.159 & 0.256 \\
 & 0.5 & \underline{\textcolor{blue}{0.072}} & \textcolor{red}{\textbf{0.168}} & 0.118 & 0.224 & 0.110 & 0.225 & 0.128 & 0.242 & 0.137 & 0.262 & 0.112 & 0.223 & 0.265 & 0.371 & 0.163 & 0.275 & \textcolor{red}{\textbf{0.072}} & \underline{\textcolor{blue}{0.173}} & 0.200 & 0.305 & 0.169 & 0.262 \\
 & 0.7 & \textcolor{red}{\textbf{0.081}} & \textcolor{red}{\textbf{0.180}} & 0.165 & 0.267 & 0.118 & 0.233 & 0.161 & 0.275 & 0.170 & 0.297 & 0.154 & 0.260 & 0.346 & 0.433 & 0.215 & 0.321 & \underline{\textcolor{blue}{0.084}} & \underline{\textcolor{blue}{0.187}} & 0.212 & 0.312 & 0.211 & 0.304 \\
 & Avg & \textcolor{red}{\textbf{0.065}} & \textcolor{red}{\textbf{0.160}} & 0.109 & 0.210 & 0.104 & 0.220 & 0.117 & 0.229 & 0.131 & 0.253 & 0.104 & 0.212 & 0.233 & 0.341 & 0.149 & 0.259 & \underline{\textcolor{blue}{0.066}} & \underline{\textcolor{blue}{0.163}} & 0.197 & 0.304 & 0.170 & 0.266 \\
\bottomrule
\end{tabular}
}
\label{tab:block_full}
\end{table*}

%% file: Appendix/Tables/random_robust_full.tex
\begin{table*}[t]
\centering
\scriptsize
\setlength{\tabcolsep}{2pt}
\caption{Full results under random missing-rate shifts across datasets, with models trained at a missing rate of $0.3$.}
\resizebox{\textwidth}{!}{%
\begin{tabular}{ll|rr|rr|rr|rr|rr|rr|rr|rr|rr|rr}
\toprule
\multicolumn{2}{c|}{Missing Ratio} & \multicolumn{2}{c|}{\textbf{GLAIM (Ours)}} & \multicolumn{2}{c|}{TimeDART} & \multicolumn{2}{c|}{ModernTCN} & \multicolumn{2}{c|}{iTransformer} & \multicolumn{2}{c|}{TimesNet} & \multicolumn{2}{c|}{PatchTST} & \multicolumn{2}{c|}{DLinear} & \multicolumn{2}{c|}{FreTS} & \multicolumn{2}{c|}{ImputeFormer} & \multicolumn{2}{c}{SAITS} \\
\multicolumn{2}{c|}{Dataset} & MSE & MAE & MSE & MAE & MSE & MAE & MSE & MAE & MSE & MAE & MSE & MAE & MSE & MAE & MSE & MAE & MSE & MAE & MSE & MAE \\
\midrule
\multirow{10}{*}{0.1} & ETTh1 & \underline{\textcolor{blue}{0.024}} & \underline{\textcolor{blue}{0.104}} & 0.092 & 0.200 & 0.037 & 0.134 & 0.100 & 0.230 & 0.077 & 0.189 & 0.079 & 0.187 & 0.171 & 0.308 & 0.119 & 0.242 & 0.041 & 0.135 & \textcolor{red}{\textbf{0.022}} & \textcolor{red}{\textbf{0.100}} \\
 & ETTh2 & \textcolor{red}{\textbf{0.027}} & \textcolor{red}{\textbf{0.095}} & 0.060 & 0.154 & \underline{\textcolor{blue}{0.039}} & \underline{\textcolor{blue}{0.126}} & 0.257 & 0.359 & 0.044 & 0.140 & 0.056 & 0.147 & 0.311 & 0.398 & 0.107 & 0.219 & 0.065 & 0.144 & 0.069 & 0.173 \\
 & ETTm1 & \underline{\textcolor{blue}{0.013}} & \textcolor{red}{\textbf{0.072}} & 0.037 & 0.122 & 0.017 & 0.086 & 0.054 & 0.163 & 0.024 & 0.103 & 0.039 & 0.127 & 0.122 & 0.260 & 0.057 & 0.164 & 0.016 & 0.083 & \textcolor{red}{\textbf{0.013}} & \underline{\textcolor{blue}{0.076}} \\
 & ETTm2 & \textcolor{red}{\textbf{0.012}} & \textcolor{red}{\textbf{0.055}} & 0.026 & 0.096 & \underline{\textcolor{blue}{0.017}} & \underline{\textcolor{blue}{0.076}} & 0.119 & 0.222 & 0.019 & 0.083 & 0.025 & 0.093 & 0.292 & 0.390 & 0.045 & 0.140 & 0.027 & 0.085 & 0.023 & 0.095 \\
 & Weather & \textcolor{red}{\textbf{0.022}} & \textcolor{red}{\textbf{0.033}} & 0.027 & 0.044 & \underline{\textcolor{blue}{0.024}} & \underline{\textcolor{blue}{0.041}} & 0.040 & 0.096 & 0.029 & 0.062 & 0.029 & 0.049 & 0.081 & 0.186 & 0.043 & 0.102 & 0.034 & 0.065 & 0.032 & 0.073 \\
 & PEMS03 & \textcolor{red}{\textbf{0.013}} & \textcolor{red}{\textbf{0.073}} & 0.032 & 0.124 & 0.033 & 0.127 & 0.043 & 0.146 & 0.041 & 0.142 & 0.037 & 0.137 & 0.115 & 0.274 & 0.039 & 0.137 & \underline{\textcolor{blue}{0.019}} & \underline{\textcolor{blue}{0.087}} & 0.062 & 0.146 \\
 & Exchange & \textcolor{red}{\textbf{0.002}} & \textcolor{red}{\textbf{0.017}} & 0.003 & 0.033 & 0.003 & \underline{\textcolor{blue}{0.027}} & 0.045 & 0.135 & 0.003 & 0.032 & \underline{\textcolor{blue}{0.003}} & 0.027 & 0.269 & 0.414 & 0.032 & 0.119 & 0.021 & 0.077 & 0.020 & 0.101 \\
 & Illness & \textcolor{red}{\textbf{0.017}} & \textcolor{red}{\textbf{0.071}} & 0.137 & 0.237 & 0.177 & 0.300 & 0.406 & 0.448 & 0.172 & 0.259 & \underline{\textcolor{blue}{0.133}} & \underline{\textcolor{blue}{0.227}} & 0.558 & 0.494 & 0.389 & 0.406 & 0.602 & 0.438 & 0.271 & 0.280 \\
 & Electricity & \textcolor{red}{\textbf{0.033}} & \textcolor{red}{\textbf{0.117}} & 0.049 & 0.152 & 0.088 & 0.212 & 0.074 & 0.190 & 0.093 & 0.207 & 0.052 & 0.158 & 0.140 & 0.285 & 0.074 & 0.187 & \underline{\textcolor{blue}{0.037}} & \underline{\textcolor{blue}{0.122}} & 0.159 & 0.276 \\
 & Avg & \textcolor{red}{\textbf{0.018}} & \textcolor{red}{\textbf{0.071}} & 0.051 & 0.129 & \underline{\textcolor{blue}{0.048}} & \underline{\textcolor{blue}{0.125}} & 0.126 & 0.221 & 0.056 & 0.135 & 0.050 & 0.128 & 0.229 & 0.334 & 0.101 & 0.191 & 0.096 & 0.137 & 0.074 & 0.147 \\
\midrule
\multirow{10}{*}{0.3} & ETTh1 & \underline{\textcolor{blue}{0.035}} & \underline{\textcolor{blue}{0.122}} & 0.116 & 0.223 & 0.046 & 0.145 & 0.118 & 0.243 & 0.074 & 0.185 & 0.089 & 0.196 & 0.169 & 0.287 & 0.140 & 0.259 & 0.060 & 0.155 & \textcolor{red}{\textbf{0.033}} & \textcolor{red}{\textbf{0.118}} \\
 & ETTh2 & \textcolor{red}{\textbf{0.031}} & \textcolor{red}{\textbf{0.104}} & 0.067 & 0.167 & \underline{\textcolor{blue}{0.041}} & \underline{\textcolor{blue}{0.130}} & 0.112 & 0.230 & 0.049 & 0.148 & 0.060 & 0.152 & 0.169 & 0.280 & 0.110 & 0.223 & 0.071 & 0.156 & 0.087 & 0.192 \\
 & ETTm1 & \textcolor{red}{\textbf{0.016}} & \textcolor{red}{\textbf{0.080}} & 0.040 & 0.127 & 0.020 & 0.091 & 0.057 & 0.164 & 0.027 & 0.110 & 0.041 & 0.129 & 0.088 & 0.206 & 0.064 & 0.172 & 0.021 & 0.092 & \underline{\textcolor{blue}{0.017}} & \underline{\textcolor{blue}{0.089}} \\
 & ETTm2 & \textcolor{red}{\textbf{0.015}} & \textcolor{red}{\textbf{0.062}} & 0.031 & 0.105 & \underline{\textcolor{blue}{0.019}} & \underline{\textcolor{blue}{0.081}} & 0.050 & 0.146 & 0.021 & 0.088 & 0.028 & 0.097 & 0.108 & 0.220 & 0.047 & 0.139 & 0.030 & 0.092 & 0.028 & 0.106 \\
 & Weather & \textcolor{red}{\textbf{0.024}} & \textcolor{red}{\textbf{0.035}} & 0.030 & 0.047 & \underline{\textcolor{blue}{0.025}} & \underline{\textcolor{blue}{0.040}} & 0.039 & 0.084 & 0.029 & 0.055 & 0.031 & 0.051 & 0.051 & 0.115 & 0.043 & 0.095 & 0.036 & 0.065 & 0.034 & 0.073 \\
 & PEMS03 & \textcolor{red}{\textbf{0.015}} & \textcolor{red}{\textbf{0.078}} & 0.034 & 0.126 & 0.028 & 0.113 & 0.038 & 0.133 & 0.036 & 0.130 & 0.038 & 0.138 & 0.063 & 0.185 & 0.040 & 0.139 & \underline{\textcolor{blue}{0.021}} & \underline{\textcolor{blue}{0.092}} & 0.061 & 0.144 \\
 & Exchange & \textcolor{red}{\textbf{0.002}} & \textcolor{red}{\textbf{0.019}} & 0.003 & 0.035 & \underline{\textcolor{blue}{0.003}} & \underline{\textcolor{blue}{0.028}} & 0.010 & 0.066 & 0.003 & 0.032 & 0.003 & 0.028 & 0.046 & 0.148 & 0.012 & 0.075 & 0.022 & 0.079 & 0.022 & 0.106 \\
 & Illness & \textcolor{red}{\textbf{0.026}} & \textcolor{red}{\textbf{0.085}} & 0.175 & 0.255 & 0.178 & 0.257 & 0.681 & 0.575 & 0.164 & 0.252 & \underline{\textcolor{blue}{0.153}} & \underline{\textcolor{blue}{0.239}} & 0.471 & 0.410 & 0.453 & 0.415 & 0.699 & 0.479 & 0.341 & 0.315 \\
 & Electricity & \textcolor{red}{\textbf{0.038}} & \textcolor{red}{\textbf{0.124}} & 0.056 & 0.158 & 0.074 & 0.188 & 0.075 & 0.188 & 0.097 & 0.213 & 0.059 & 0.165 & 0.125 & 0.256 & 0.088 & 0.204 & \underline{\textcolor{blue}{0.040}} & \underline{\textcolor{blue}{0.128}} & 0.163 & 0.278 \\
 & Avg & \textcolor{red}{\textbf{0.022}} & \textcolor{red}{\textbf{0.079}} & 0.061 & 0.138 & \underline{\textcolor{blue}{0.048}} & \underline{\textcolor{blue}{0.119}} & 0.131 & 0.203 & 0.056 & 0.135 & 0.056 & 0.133 & 0.143 & 0.234 & 0.111 & 0.191 & 0.111 & 0.149 & 0.087 & 0.158 \\
\midrule
\multirow{10}{*}{0.5} & ETTh1 & \textcolor{red}{\textbf{0.053}} & \textcolor{red}{\textbf{0.149}} & 0.167 & 0.261 & 0.073 & 0.179 & 0.215 & 0.335 & 0.105 & 0.215 & 0.121 & 0.223 & 0.291 & 0.375 & 0.220 & 0.326 & 0.093 & 0.188 & \underline{\textcolor{blue}{0.055}} & \underline{\textcolor{blue}{0.153}} \\
 & ETTh2 & \textcolor{red}{\textbf{0.039}} & \textcolor{red}{\textbf{0.119}} & 0.082 & 0.186 & \underline{\textcolor{blue}{0.052}} & \underline{\textcolor{blue}{0.147}} & 0.372 & 0.432 & 0.065 & 0.169 & 0.071 & 0.169 & 0.468 & 0.473 & 0.265 & 0.351 & 0.086 & 0.176 & 0.120 & 0.222 \\
 & ETTm1 & \textcolor{red}{\textbf{0.022}} & \textcolor{red}{\textbf{0.093}} & 0.051 & 0.140 & 0.033 & 0.115 & 0.178 & 0.292 & 0.044 & 0.135 & 0.055 & 0.145 & 0.198 & 0.309 & 0.117 & 0.238 & 0.029 & 0.107 & \underline{\textcolor{blue}{0.026}} & \underline{\textcolor{blue}{0.107}} \\
 & ETTm2 & \textcolor{red}{\textbf{0.019}} & \textcolor{red}{\textbf{0.073}} & 0.039 & 0.120 & \underline{\textcolor{blue}{0.025}} & \underline{\textcolor{blue}{0.096}} & 0.722 & 0.534 & 0.028 & 0.102 & 0.034 & 0.108 & 0.390 & 0.433 & 0.126 & 0.234 & 0.037 & 0.106 & 0.039 & 0.125 \\
 & Weather & \textcolor{red}{\textbf{0.028}} & \textcolor{red}{\textbf{0.039}} & 0.036 & 0.058 & \underline{\textcolor{blue}{0.031}} & \underline{\textcolor{blue}{0.052}} & 0.118 & 0.217 & 0.037 & 0.071 & 0.037 & 0.061 & 0.108 & 0.206 & 0.079 & 0.169 & 0.039 & 0.068 & 0.077 & 0.109 \\
 & PEMS03 & \textcolor{red}{\textbf{0.019}} & \textcolor{red}{\textbf{0.089}} & 0.037 & 0.132 & 0.046 & 0.154 & 0.117 & 0.268 & 0.061 & 0.181 & 0.043 & 0.145 & 0.153 & 0.306 & 0.062 & 0.179 & \underline{\textcolor{blue}{0.025}} & \underline{\textcolor{blue}{0.101}} & 0.105 & 0.235 \\
 & Exchange & \textcolor{red}{\textbf{0.002}} & \textcolor{red}{\textbf{0.023}} & 0.005 & 0.043 & 0.004 & 0.036 & 1.170 & 0.696 & 0.004 & 0.037 & \underline{\textcolor{blue}{0.003}} & \underline{\textcolor{blue}{0.033}} & 0.335 & 0.446 & 0.065 & 0.170 & 0.024 & 0.087 & 0.029 & 0.122 \\
 & Illness & \textcolor{red}{\textbf{0.042}} & \textcolor{red}{\textbf{0.108}} & 0.296 & 0.317 & 0.387 & 0.403 & 1.197 & 0.753 & \underline{\textcolor{blue}{0.218}} & \underline{\textcolor{blue}{0.283}} & 0.260 & 0.295 & 0.998 & 0.583 & 0.724 & 0.511 & 0.896 & 0.557 & 0.479 & 0.382 \\
 & Electricity & \textcolor{red}{\textbf{0.047}} & \textcolor{red}{\textbf{0.139}} & 0.073 & 0.176 & 0.088 & 0.212 & 0.143 & 0.272 & 0.106 & 0.227 & 0.082 & 0.190 & 0.230 & 0.356 & 0.148 & 0.272 & \underline{\textcolor{blue}{0.048}} & \underline{\textcolor{blue}{0.139}} & 0.181 & 0.297 \\
 & Avg & \textcolor{red}{\textbf{0.030}} & \textcolor{red}{\textbf{0.093}} & 0.087 & 0.159 & 0.082 & 0.155 & 0.470 & 0.422 & \underline{\textcolor{blue}{0.074}} & 0.158 & 0.078 & \underline{\textcolor{blue}{0.152}} & 0.352 & 0.388 & 0.201 & 0.272 & 0.142 & 0.170 & 0.123 & 0.195 \\
\midrule
\multirow{10}{*}{0.7} & ETTh1 & \textcolor{red}{\textbf{0.112}} & \textcolor{red}{\textbf{0.210}} & 0.274 & 0.330 & 0.168 & 0.262 & 0.464 & 0.511 & 0.224 & 0.303 & 0.228 & 0.296 & 0.532 & 0.524 & 0.435 & 0.470 & 0.165 & 0.249 & \underline{\textcolor{blue}{0.123}} & \underline{\textcolor{blue}{0.229}} \\
 & ETTh2 & \textcolor{red}{\textbf{0.059}} & \textcolor{red}{\textbf{0.150}} & 0.110 & 0.218 & \underline{\textcolor{blue}{0.082}} & \underline{\textcolor{blue}{0.187}} & 1.085 & 0.778 & 0.101 & 0.213 & 0.099 & 0.203 & 1.208 & 0.809 & 0.999 & 0.729 & 0.116 & 0.212 & 0.217 & 0.299 \\
 & ETTm1 & \textcolor{red}{\textbf{0.040}} & \textcolor{red}{\textbf{0.123}} & 0.100 & 0.191 & 0.108 & 0.196 & 0.563 & 0.560 & 0.125 & 0.216 & 0.125 & 0.206 & 0.454 & 0.488 & 0.307 & 0.393 & 0.062 & \underline{\textcolor{blue}{0.148}} & \underline{\textcolor{blue}{0.055}} & 0.150 \\
 & ETTm2 & \textcolor{red}{\textbf{0.028}} & \textcolor{red}{\textbf{0.091}} & 0.056 & 0.148 & \underline{\textcolor{blue}{0.042}} & \underline{\textcolor{blue}{0.127}} & 2.593 & 1.197 & 0.046 & 0.134 & 0.050 & 0.136 & 1.135 & 0.793 & 0.769 & 0.618 & 0.063 & 0.142 & 0.097 & 0.191 \\
 & Weather & \textcolor{red}{\textbf{0.036}} & \textcolor{red}{\textbf{0.050}} & 0.053 & 0.089 & \underline{\textcolor{blue}{0.047}} & \underline{\textcolor{blue}{0.085}} & 0.352 & 0.441 & 0.058 & 0.105 & 0.054 & 0.091 & 0.253 & 0.362 & 0.221 & 0.339 & 0.057 & 0.100 & 0.851 & 0.290 \\
 & PEMS03 & \textcolor{red}{\textbf{0.029}} & \textcolor{red}{\textbf{0.115}} & 0.046 & 0.146 & 0.137 & 0.275 & 0.583 & 0.640 & 0.155 & 0.293 & 0.073 & 0.185 & 0.386 & 0.516 & 0.199 & 0.349 & \underline{\textcolor{blue}{0.035}} & \underline{\textcolor{blue}{0.122}} & 0.296 & 0.434 \\
 & Exchange & \textcolor{red}{\textbf{0.003}} & \textcolor{red}{\textbf{0.031}} & 0.009 & 0.059 & 0.008 & 0.055 & 3.170 & 1.465 & 0.007 & 0.053 & \underline{\textcolor{blue}{0.007}} & \underline{\textcolor{blue}{0.048}} & 1.140 & 0.869 & 0.687 & 0.631 & 0.043 & 0.132 & 0.047 & 0.157 \\
 & Illness & \textcolor{red}{\textbf{0.107}} & \textcolor{red}{\textbf{0.175}} & 0.646 & 0.483 & 0.963 & 0.685 & 2.213 & 1.026 & \underline{\textcolor{blue}{0.474}} & \underline{\textcolor{blue}{0.449}} & 0.604 & 0.460 & 2.129 & 0.928 & 1.448 & 0.751 & 1.458 & 0.731 & 0.854 & 0.534 \\
 & Electricity & \textcolor{red}{\textbf{0.075}} & \underline{\textcolor{blue}{0.182}} & 0.134 & 0.242 & 0.184 & 0.330 & 0.414 & 0.497 & 0.174 & 0.299 & 0.175 & 0.286 & 0.452 & 0.529 & 0.335 & 0.439 & \underline{\textcolor{blue}{0.076}} & \textcolor{red}{\textbf{0.178}} & 0.254 & 0.369 \\
 & Avg & \textcolor{red}{\textbf{0.054}} & \textcolor{red}{\textbf{0.125}} & 0.159 & \underline{\textcolor{blue}{0.212}} & 0.193 & 0.245 & 1.271 & 0.791 & \underline{\textcolor{blue}{0.151}} & 0.229 & 0.157 & 0.212 & 0.854 & 0.646 & 0.600 & 0.524 & 0.231 & 0.224 & 0.310 & 0.295 \\
\bottomrule
\end{tabular}
}
\label{tab:random_robust}
\end{table*}

%% file: Appendix/Tables/multi_block_robust_full.tex
\begin{table*}[t]
\centering
\scriptsize
\setlength{\tabcolsep}{2pt}
\caption{Full results under block missing-rate shifts across datasets, with models trained at a missing rate of $0.3$.}
\resizebox{\textwidth}{!}{%
\begin{tabular}{ll|rr|rr|rr|rr|rr|rr|rr|rr|rr|rr}
\toprule
\multicolumn{2}{c|}{Missing Ratio} & \multicolumn{2}{c|}{\textbf{GLAIM (Ours)}} & \multicolumn{2}{c|}{TimeDART} & \multicolumn{2}{c|}{ModernTCN} & \multicolumn{2}{c|}{iTransformer} & \multicolumn{2}{c|}{TimesNet} & \multicolumn{2}{c|}{PatchTST} & \multicolumn{2}{c|}{DLinear} & \multicolumn{2}{c|}{FreTS} & \multicolumn{2}{c|}{ImputeFormer} & \multicolumn{2}{c}{SAITS} \\
\multicolumn{2}{c|}{Dataset} & MSE & MAE & MSE & MAE & MSE & MAE & MSE & MAE & MSE & MAE & MSE & MAE & MSE & MAE & MSE & MAE & MSE & MAE & MSE & MAE \\
\midrule
\multirow{10}{*}{0.1} & ETTh1 & \underline{\textcolor{blue}{0.033}} & \underline{\textcolor{blue}{0.125}} & 0.185 & 0.284 & 0.063 & 0.178 & 0.194 & 0.317 & 0.115 & 0.231 & 0.136 & 0.249 & 0.301 & 0.401 & 0.262 & 0.369 & 0.087 & 0.191 & \textcolor{red}{\textbf{0.032}} & \textcolor{red}{\textbf{0.119}} \\
 & ETTh2 & \textcolor{red}{\textbf{0.033}} & \textcolor{red}{\textbf{0.110}} & 0.083 & 0.191 & \underline{\textcolor{blue}{0.055}} & \underline{\textcolor{blue}{0.157}} & 0.222 & 0.334 & 0.058 & 0.162 & 0.077 & 0.183 & 0.552 & 0.545 & 0.317 & 0.408 & 0.115 & 0.205 & 0.143 & 0.262 \\
 & ETTm1 & \underline{\textcolor{blue}{0.016}} & \underline{\textcolor{blue}{0.085}} & 0.064 & 0.166 & 0.025 & 0.110 & 0.159 & 0.282 & 0.036 & 0.129 & 0.056 & 0.154 & 0.218 & 0.351 & 0.127 & 0.262 & 0.024 & 0.105 & \textcolor{red}{\textbf{0.016}} & \textcolor{red}{\textbf{0.085}} \\
 & ETTm2 & \textcolor{red}{\textbf{0.016}} & \textcolor{red}{\textbf{0.065}} & 0.044 & 0.131 & \underline{\textcolor{blue}{0.025}} & \underline{\textcolor{blue}{0.097}} & 0.284 & 0.341 & 0.028 & 0.105 & 0.039 & 0.120 & 0.441 & 0.493 & 0.185 & 0.304 & 0.046 & 0.121 & 0.030 & 0.113 \\
 & Weather & \textcolor{red}{\textbf{0.029}} & \textcolor{red}{\textbf{0.042}} & 0.039 & 0.057 & \underline{\textcolor{blue}{0.034}} & \underline{\textcolor{blue}{0.057}} & 0.071 & 0.134 & 0.039 & 0.070 & 0.043 & 0.064 & 0.127 & 0.235 & 0.083 & 0.169 & 0.046 & 0.078 & 0.049 & 0.103 \\
 & PEMS03 & \textcolor{red}{\textbf{0.015}} & \textcolor{red}{\textbf{0.079}} & 0.041 & 0.141 & 0.039 & 0.135 & 0.063 & 0.174 & 0.055 & 0.169 & 0.044 & 0.148 & 0.154 & 0.317 & 0.061 & 0.170 & \underline{\textcolor{blue}{0.021}} & \underline{\textcolor{blue}{0.092}} & 0.063 & 0.146 \\
 & Exchange & \textcolor{red}{\textbf{0.003}} & \textcolor{red}{\textbf{0.025}} & 0.005 & 0.044 & 0.005 & 0.044 & 0.189 & 0.251 & 0.007 & 0.048 & \underline{\textcolor{blue}{0.004}} & \underline{\textcolor{blue}{0.038}} & 0.318 & 0.456 & 0.143 & 0.293 & 0.028 & 0.094 & 0.114 & 0.291 \\
 & Illness & \textcolor{red}{\textbf{0.068}} & \textcolor{red}{\textbf{0.142}} & 0.425 & 0.390 & \underline{\textcolor{blue}{0.281}} & \underline{\textcolor{blue}{0.355}} & 0.593 & 0.517 & 0.356 & 0.368 & 0.454 & 0.395 & 0.940 & 0.694 & 1.308 & 0.689 & 0.760 & 0.519 & 0.459 & 0.413 \\
 & Electricity & \textcolor{red}{\textbf{0.049}} & \textcolor{red}{\textbf{0.143}} & 0.070 & 0.174 & 0.132 & 0.260 & 0.107 & 0.226 & 0.102 & 0.216 & 0.070 & 0.181 & 0.188 & 0.315 & 0.117 & 0.236 & \underline{\textcolor{blue}{0.052}} & \underline{\textcolor{blue}{0.146}} & 0.192 & 0.304 \\
 & Avg & \textcolor{red}{\textbf{0.029}} & \textcolor{red}{\textbf{0.091}} & 0.106 & 0.175 & \underline{\textcolor{blue}{0.073}} & \underline{\textcolor{blue}{0.155}} & 0.209 & 0.286 & 0.088 & 0.166 & 0.103 & 0.170 & 0.360 & 0.423 & 0.289 & 0.322 & 0.131 & 0.172 & 0.122 & 0.204 \\
\midrule
\multirow{10}{*}{0.3} & ETTh1 & \textcolor{red}{\textbf{0.054}} & \textcolor{red}{\textbf{0.150}} & 0.237 & 0.315 & 0.090 & 0.200 & 0.217 & 0.319 & 0.132 & 0.241 & 0.188 & 0.281 & 0.317 & 0.381 & 0.277 & 0.354 & 0.151 & 0.234 & \underline{\textcolor{blue}{0.069}} & \underline{\textcolor{blue}{0.159}} \\
 & ETTh2 & \textcolor{red}{\textbf{0.043}} & \textcolor{red}{\textbf{0.128}} & 0.107 & 0.218 & \underline{\textcolor{blue}{0.061}} & \underline{\textcolor{blue}{0.166}} & 0.143 & 0.264 & 0.071 & 0.179 & 0.092 & 0.198 & 0.232 & 0.336 & 0.142 & 0.259 & 0.127 & 0.223 & 0.181 & 0.298 \\
 & ETTm1 & \textcolor{red}{\textbf{0.023}} & \textcolor{red}{\textbf{0.096}} & 0.088 & 0.192 & 0.034 & 0.123 & 0.119 & 0.235 & 0.051 & 0.151 & 0.075 & 0.175 & 0.240 & 0.333 & 0.141 & 0.249 & 0.036 & 0.123 & \underline{\textcolor{blue}{0.028}} & \underline{\textcolor{blue}{0.108}} \\
 & ETTm2 & \textcolor{red}{\textbf{0.021}} & \textcolor{red}{\textbf{0.076}} & 0.055 & 0.147 & \underline{\textcolor{blue}{0.029}} & \underline{\textcolor{blue}{0.105}} & 0.070 & 0.175 & 0.033 & 0.116 & 0.049 & 0.134 & 0.141 & 0.262 & 0.068 & 0.173 & 0.055 & 0.135 & 0.039 & 0.131 \\
 & Weather & \textcolor{red}{\textbf{0.035}} & \textcolor{red}{\textbf{0.048}} & 0.049 & 0.066 & \underline{\textcolor{blue}{0.038}} & \underline{\textcolor{blue}{0.058}} & 0.065 & 0.110 & 0.045 & 0.075 & 0.052 & 0.072 & 0.085 & 0.154 & 0.064 & 0.112 & 0.050 & 0.081 & 0.059 & 0.117 \\
 & PEMS03 & \textcolor{red}{\textbf{0.017}} & \textcolor{red}{\textbf{0.081}} & 0.046 & 0.145 & 0.037 & 0.128 & 0.049 & 0.147 & 0.051 & 0.157 & 0.048 & 0.150 & 0.109 & 0.242 & 0.057 & 0.162 & \underline{\textcolor{blue}{0.025}} & \underline{\textcolor{blue}{0.098}} & 0.063 & 0.144 \\
 & Exchange & \textcolor{red}{\textbf{0.004}} & \textcolor{red}{\textbf{0.033}} & 0.006 & 0.050 & 0.006 & 0.050 & 0.012 & 0.073 & 0.008 & 0.054 & \underline{\textcolor{blue}{0.006}} & \underline{\textcolor{blue}{0.046}} & 0.039 & 0.136 & 0.012 & 0.075 & 0.030 & 0.102 & 0.128 & 0.301 \\
 & Illness & \textcolor{red}{\textbf{0.093}} & \textcolor{red}{\textbf{0.160}} & 0.656 & 0.492 & 0.419 & 0.405 & 0.876 & 0.617 & \underline{\textcolor{blue}{0.404}} & \underline{\textcolor{blue}{0.392}} & 0.650 & 0.483 & 1.223 & 0.699 & 2.028 & 0.880 & 1.065 & 0.620 & 0.635 & 0.498 \\
 & Electricity & \textcolor{red}{\textbf{0.060}} & \textcolor{red}{\textbf{0.155}} & 0.089 & 0.192 & 0.104 & 0.220 & 0.104 & 0.217 & 0.115 & 0.236 & 0.086 & 0.195 & 0.197 & 0.313 & 0.129 & 0.241 & \underline{\textcolor{blue}{0.061}} & \underline{\textcolor{blue}{0.157}} & 0.194 & 0.305 \\
 & Avg & \textcolor{red}{\textbf{0.039}} & \textcolor{red}{\textbf{0.103}} & 0.148 & 0.202 & \underline{\textcolor{blue}{0.091}} & \underline{\textcolor{blue}{0.162}} & 0.184 & 0.240 & 0.101 & 0.178 & 0.138 & 0.193 & 0.287 & 0.317 & 0.324 & 0.278 & 0.178 & 0.197 & 0.155 & 0.229 \\
\midrule
\multirow{10}{*}{0.5} & ETTh1 & \textcolor{red}{\textbf{0.097}} & \textcolor{red}{\textbf{0.200}} & 0.305 & 0.359 & 0.148 & 0.253 & 0.313 & 0.397 & 0.214 & 0.298 & 0.245 & 0.320 & 0.418 & 0.443 & 0.355 & 0.415 & 0.225 & 0.287 & \underline{\textcolor{blue}{0.128}} & \underline{\textcolor{blue}{0.213}} \\
 & ETTh2 & \textcolor{red}{\textbf{0.058}} & \textcolor{red}{\textbf{0.152}} & 0.122 & 0.233 & \underline{\textcolor{blue}{0.076}} & \underline{\textcolor{blue}{0.186}} & 0.383 & 0.437 & 0.090 & 0.202 & 0.105 & 0.213 & 0.486 & 0.489 & 0.397 & 0.434 & 0.147 & 0.247 & 0.212 & 0.326 \\
 & ETTm1 & \textcolor{red}{\textbf{0.039}} & \textcolor{red}{\textbf{0.126}} & 0.120 & 0.217 & 0.067 & 0.163 & 0.198 & 0.322 & 0.082 & 0.184 & 0.101 & 0.199 & 0.340 & 0.398 & 0.211 & 0.316 & 0.052 & 0.144 & \underline{\textcolor{blue}{0.048}} & \underline{\textcolor{blue}{0.141}} \\
 & ETTm2 & \textcolor{red}{\textbf{0.028}} & \textcolor{red}{\textbf{0.093}} & 0.062 & 0.156 & \underline{\textcolor{blue}{0.037}} & \underline{\textcolor{blue}{0.121}} & 0.420 & 0.446 & 0.042 & 0.130 & 0.056 & 0.145 & 0.397 & 0.444 & 0.241 & 0.335 & 0.063 & 0.148 & 0.053 & 0.155 \\
 & Weather & \textcolor{red}{\textbf{0.041}} & \textcolor{red}{\textbf{0.057}} & 0.054 & 0.076 & \underline{\textcolor{blue}{0.044}} & \underline{\textcolor{blue}{0.070}} & 0.115 & 0.207 & 0.055 & 0.093 & 0.059 & 0.087 & 0.136 & 0.228 & 0.111 & 0.202 & 0.056 & 0.088 & 0.089 & 0.171 \\
 & PEMS03 & \textcolor{red}{\textbf{0.024}} & \textcolor{red}{\textbf{0.104}} & 0.053 & 0.155 & 0.052 & 0.159 & 0.224 & 0.363 & 0.080 & 0.202 & 0.055 & 0.159 & 0.196 & 0.339 & 0.090 & 0.211 & \underline{\textcolor{blue}{0.030}} & \underline{\textcolor{blue}{0.109}} & 0.106 & 0.234 \\
 & Exchange & \textcolor{red}{\textbf{0.005}} & \textcolor{red}{\textbf{0.039}} & 0.007 & 0.054 & 0.008 & 0.055 & 0.802 & 0.656 & 0.009 & 0.060 & \underline{\textcolor{blue}{0.007}} & \underline{\textcolor{blue}{0.051}} & 0.339 & 0.456 & 0.203 & 0.317 & 0.033 & 0.111 & 0.145 & 0.314 \\
 & Illness & \textcolor{red}{\textbf{0.193}} & \textcolor{red}{\textbf{0.246}} & 0.939 & 0.595 & 0.756 & 0.554 & 1.346 & 0.750 & \underline{\textcolor{blue}{0.607}} & \underline{\textcolor{blue}{0.483}} & 0.910 & 0.578 & 1.688 & 0.798 & 2.490 & 0.984 & 1.617 & 0.776 & 1.002 & 0.631 \\
 & Electricity & \underline{\textcolor{blue}{0.072}} & \underline{\textcolor{blue}{0.174}} & 0.116 & 0.220 & 0.131 & 0.263 & 0.176 & 0.301 & 0.139 & 0.265 & 0.119 & 0.227 & 0.305 & 0.406 & 0.212 & 0.327 & \textcolor{red}{\textbf{0.070}} & \textcolor{red}{\textbf{0.169}} & 0.214 & 0.329 \\
 & Avg & \textcolor{red}{\textbf{0.062}} & \textcolor{red}{\textbf{0.132}} & 0.198 & 0.230 & 0.147 & \underline{\textcolor{blue}{0.203}} & 0.442 & 0.431 & \underline{\textcolor{blue}{0.146}} & 0.213 & 0.184 & 0.220 & 0.478 & 0.445 & 0.479 & 0.394 & 0.255 & 0.231 & 0.222 & 0.279 \\
\midrule
\multirow{10}{*}{0.7} & ETTh1 & \textcolor{red}{\textbf{0.217}} & \textcolor{red}{\textbf{0.291}} & 0.417 & 0.420 & 0.267 & 0.337 & 0.491 & 0.515 & 0.364 & 0.385 & 0.392 & 0.402 & 0.587 & 0.544 & 0.513 & 0.515 & 0.336 & 0.361 & \underline{\textcolor{blue}{0.237}} & \underline{\textcolor{blue}{0.294}} \\
 & ETTh2 & \textcolor{red}{\textbf{0.092}} & \textcolor{red}{\textbf{0.194}} & 0.142 & 0.251 & \underline{\textcolor{blue}{0.104}} & \underline{\textcolor{blue}{0.217}} & 1.187 & 0.804 & 0.117 & 0.229 & 0.128 & 0.236 & 1.114 & 0.773 & 0.869 & 0.650 & 0.174 & 0.277 & 0.252 & 0.360 \\
 & ETTm1 & \textcolor{red}{\textbf{0.078}} & \underline{\textcolor{blue}{0.180}} & 0.193 & 0.268 & 0.137 & 0.226 & 0.427 & 0.484 & 0.141 & 0.233 & 0.172 & 0.252 & 0.503 & 0.505 & 0.321 & 0.392 & \underline{\textcolor{blue}{0.078}} & \textcolor{red}{\textbf{0.174}} & 0.083 & 0.186 \\
 & ETTm2 & \textcolor{red}{\textbf{0.043}} & \textcolor{red}{\textbf{0.122}} & 0.072 & 0.170 & \underline{\textcolor{blue}{0.052}} & \underline{\textcolor{blue}{0.144}} & 1.741 & 0.977 & 0.054 & 0.148 & 0.067 & 0.163 & 1.059 & 0.764 & 0.732 & 0.599 & 0.076 & 0.166 & 0.081 & 0.193 \\
 & Weather & \textcolor{red}{\textbf{0.051}} & \textcolor{red}{\textbf{0.073}} & 0.065 & 0.097 & \underline{\textcolor{blue}{0.056}} & \underline{\textcolor{blue}{0.091}} & 0.269 & 0.372 & 0.074 & 0.121 & 0.074 & 0.115 & 0.248 & 0.351 & 0.226 & 0.332 & 0.067 & 0.103 & 0.199 & 0.303 \\
 & PEMS03 & \underline{\textcolor{blue}{0.049}} & \underline{\textcolor{blue}{0.159}} & 0.071 & 0.181 & 0.121 & 0.255 & 0.747 & 0.712 & 0.174 & 0.304 & 0.097 & 0.214 & 0.381 & 0.503 & 0.189 & 0.325 & \textcolor{red}{\textbf{0.043}} & \textcolor{red}{\textbf{0.132}} & 0.280 & 0.418 \\
 & Exchange & \textcolor{red}{\textbf{0.007}} & \textcolor{red}{\textbf{0.048}} & 0.010 & 0.063 & 0.010 & 0.064 & 3.109 & 1.448 & 0.011 & 0.067 & \underline{\textcolor{blue}{0.009}} & \underline{\textcolor{blue}{0.059}} & 1.139 & 0.867 & 0.668 & 0.596 & 0.043 & 0.129 & 0.202 & 0.361 \\
 & Illness & \textcolor{red}{\textbf{0.473}} & \textcolor{red}{\textbf{0.426}} & 1.323 & 0.736 & 1.379 & 0.781 & 2.094 & 0.952 & \underline{\textcolor{blue}{1.097}} & \underline{\textcolor{blue}{0.650}} & 1.270 & 0.712 & 2.487 & 0.985 & 2.751 & 1.063 & 2.394 & 0.979 & 1.649 & 0.820 \\
 & Electricity & \underline{\textcolor{blue}{0.108}} & \underline{\textcolor{blue}{0.224}} & 0.221 & 0.321 & 0.279 & 0.415 & 0.387 & 0.483 & 0.226 & 0.343 & 0.266 & 0.365 & 0.490 & 0.542 & 0.379 & 0.462 & \textcolor{red}{\textbf{0.101}} & \textcolor{red}{\textbf{0.208}} & 0.322 & 0.428 \\
 & Avg & \textcolor{red}{\textbf{0.124}} & \textcolor{red}{\textbf{0.191}} & 0.279 & 0.278 & 0.267 & 0.281 & 1.161 & 0.750 & \underline{\textcolor{blue}{0.251}} & \underline{\textcolor{blue}{0.275}} & 0.275 & 0.280 & 0.890 & 0.648 & 0.739 & 0.548 & 0.368 & 0.281 & 0.367 & 0.374 \\
\bottomrule
\end{tabular}
}
\label{tab:block_robust}
\end{table*}